\documentclass{article}
\usepackage{authblk}
\usepackage[utf8]{inputenc}
\usepackage{main}
\usepackage{microtype}
\usepackage{subcaption}
\usepackage{graphicx}
\usepackage{times}
\usepackage{latexsym}
\usepackage{amsmath}
\usepackage{float}
\usepackage{footnote}
\usepackage{enumitem}
\usepackage{bm}
\usepackage{arydshln}
\usepackage{booktabs}
\usepackage{array}     
\usepackage{multicol}
\usepackage{multirow}
\usepackage{color}
\usepackage{xcolor}     
\usepackage{colortbl}
\usepackage{bbding}
\usepackage{makecell}
\usepackage{mathtools}
\usepackage{imakeidx}
\usepackage{longtable}
\usepackage{wrapfig}
\usepackage{algorithmic}
\usepackage{rotating}
\makeindex
\usepackage{arydshln}
\usepackage{lipsum}
\usepackage{natbib}
\usepackage[toc]{multitoc}
\usepackage[edges]{forest}
\usepackage[normalem]{ulem}
\definecolor{mydarkblue}{rgb}{0,0.08,0.45}
\usepackage[colorlinks=true,linkcolor=mydarkblue,citecolor=mydarkblue,filecolor=mydarkblue,urlcolor=mydarkblue]{hyperref}
\usepackage{CJKutf8}
\usepackage{awesomebox} 
\usepackage{bbding}
\usepackage[most]{tcolorbox}
\usepackage{booktabs}
\usepackage{geometry}
\definecolor{wkblue}{rgb}{0.2, 0.3, 0.6}
\definecolor{meta-color}{rgb}{0.5, 0.5, 0.5}
\usepackage{amsmath}
\usepackage{enumitem}
\usepackage{lscape} 
\usepackage{booktabs}
\usepackage{algorithm}   
\usepackage{setspace}

\usepackage{algorithmic}
\usepackage{tabularx,booktabs}
\usepackage{makecell}

\usepackage{amssymb}
\usepackage{amsfonts}

\usepackage{booktabs} 
\usepackage{geometry} 

\usepackage{multirow}

\usepackage[tikz]{bclogo}
\usepackage[framemethod=tikz]{mdframed}
\definecolor{bgblue}{RGB}{245,243,253}
\definecolor{ttblue}{RGB}{91,194,224}

\usepackage{pgfplots}
\usepackage{pgfplotstable}

\usepackage{wrapfig}
\usepackage{graphicx}

\mdfdefinestyle{mystyle}{%
  rightline=true,
  innerleftmargin=10,
  innerrightmargin=10,
  outerlinewidth=3pt,
  topline=false,
  rightline=true,
  bottomline=false,
  skipabove=\topsep,
  skipbelow=\topsep
}

\newtcolorbox{myboxi}[1][]{
  breakable,
  title=#1,
  colback=red!5,
  colbacktitle=red!5,
  coltitle=black,
  fonttitle=\bfseries,
  bottomrule=0pt,
  toprule=0pt,
  leftrule=2pt,
  rightrule=2pt,
  titlerule=0pt,
  arc=0pt,
  outer arc=0pt,
  colframe=red,
}

\newtcolorbox{myboxnote}[1][]{
  breakable,
  title=#1,
  colback=orange!0,
  colbacktitle=orange!0,
  coltitle=black,
  fonttitle=\bfseries,
  bottomrule=0pt,
  toprule=0pt,
  leftrule=2pt,
  rightrule=2pt,
  titlerule=0pt,
  arc=0pt,
  outer arc=0pt,
  colframe=orange,
}

\newtcolorbox{myboxii}[1][]{
  breakable,
  freelance,
  title=#1,
  colback=white,
  colbacktitle=white,
  coltitle=black,
  fonttitle=\bfseries,
  bottomrule=0pt,
  boxrule=0pt,
  colframe=white,
  overlay unbroken and first={
  \draw[red!75!black,line width=3pt]
    ([xshift=5pt]frame.north west) -- 
    (frame.north west) -- 
    (frame.south west);
  \draw[red!75!black,line width=3pt]
    ([xshift=-5pt]frame.north east) -- 
    (frame.north east) -- 
    (frame.south east);
  },
  overlay unbroken app={
  \draw[red!75!black,line width=3pt,line cap=rect]
    (frame.south west) -- 
    ([xshift=5pt]frame.south west);
  \draw[red!75!black,line width=3pt,line cap=rect]
    (frame.south east) -- 
    ([xshift=-5pt]frame.south east);
  },
  overlay middle and last={
  \draw[red!75!black,line width=3pt]
    (frame.north west) -- 
    (frame.south west);
  \draw[red!75!black,line width=3pt]
    (frame.north east) -- 
    (frame.south east);
  },
  overlay last app={
  \draw[red!75!black,line width=3pt,line cap=rect]
    (frame.south west) --
    ([xshift=5pt]frame.south west);
  \draw[red!75!black,line width=3pt,line cap=rect]
    (frame.south east) --
    ([xshift=-5pt]frame.south east);
  },
}

\usepackage{fontawesome5}
\usepackage{fancyhdr} 
\usepackage{blindtext} 
\usepackage{makecell}

\DeclareCaptionFont{black}{\color{black}}

\definecolor{myblue}{rgb}{0.9, 0.1, 0.94}
\definecolor{mygreen}{rgb}{0.64, 0.56, 0.88}
\definecolor{myyellow}{rgb}{0.68, 0.6, 0.1}
\definecolor{fancygreen}{rgb}{0.33, 0.68, 0.20}
\definecolor{salmon}{rgb}{0.94, 0.52, 0.49}
\definecolor{tablegreen}{rgb}{0.82, 0.94, 0.75}
\definecolor{tableblue}{rgb}{0.81, 0.90, 0.94}
\definecolor{tablered}{rgb}{0.97, 0.85, 0.85}
\definecolor{tableorange}{rgb}{0.96, 0.85, 0.81}

\newenvironment{itemize*}%
 {\leftmargini=10pt\begin{itemize}%
  \setlength{\itemsep}{0pt}%
  \setlength{\parskip}{0pt}%
  }%
 {\end{itemize}}
\newenvironment{enumerate*}%
 {\begin{enumerate}%
  \setlength{\itemsep}{0pt}%
  \setlength{\parskip}{0pt}}%
 {\end{enumerate}}

\usepackage{xcolor}
\usepackage{listings}  

\newcommand\JSONnumbervaluestyle{\color{blue}}
\newcommand\JSONstringvaluestyle{\color{red}}

\newif\ifcolonfoundonthisline

\makeatletter

\lstdefinestyle{json}
{
  showstringspaces    = false,
  keywords            = {false,true},
  alsoletter          = 0123456789.,
  morestring          = [s]{"}{"},
  stringstyle         = \ifcolonfoundonthisline\JSONstringvaluestyle\fi,
  MoreSelectCharTable =%
    \lst@DefSaveDef{`:}\colon@json{\processColon@json},
  basicstyle          = \ttfamily,
  keywordstyle        = \ttfamily\bfseries,
}

\newcommand\processColon@json{%
  \colon@json%
  \ifnum\lst@mode=\lst@Pmode%
    \global\colonfoundonthislinetrue%
  \fi
}

\lst@AddToHook{Output}{%
  \ifcolonfoundonthisline%
    \ifnum\lst@mode=\lst@Pmode%
      \def\lst@thestyle{\JSONnumbervaluestyle}%
    \fi
  \fi
  \lsthk@DetectKeywords%
}

\lst@AddToHook{EOL}%
  {\global\colonfoundonthislinefalse}

\makeatother

\usepackage{etoolbox}
\usepackage{natbib}
\usepackage{url}
\newcounter{bibcount}
\makeatletter
\patchcmd{\@lbibitem}{\item[}{\item[\hfil\stepcounter{bibcount}{[\thebibcount]}}{}{}
\renewcommand\NAT@bibsetup%
  [1]{\setlength{\leftmargin}{\bibhang}\setlength{\itemindent}{-\parindent}%
      \setlength{\itemsep}{\bibsep}\setlength{\parsep}{\z@}}
\makeatother

\definecolor{mybrown}{RGB}{128,64,0}

\definecolor{titlecolor}{HTML}{4c9cff}
\definecolor{coolblue3}{rgb}{0.91, 0.94, 0.98}

\begin{document}



\title{LIMI: Less is More for Agency}

\author[3,5]{Yang Xiao\textsuperscript{*}}
\author[1,2,5]{Mohan Jiang\textsuperscript{*}}
\author[4,2,5]{Jie Sun\textsuperscript{*}}
\author[1,2,5]{Keyu Li\textsuperscript{*}}
\author[1,5]{Jifan Lin}
\author[1,5]{Yumin Zhuang}
\author[1,5]{Ji Zeng}  
\author[1,2,5]{Shijie Xia}
\author[1,5]{Qishuo Hua}
\author[1,2,5]{Xuefeng Li} \par
\author[1,5]{Xiaojie Cai} 
\author[2]{Tongyu Wang}
\author[2]{Yue Zhang}
\author[2]{Liming Liu}
\author[2]{Xia Wu}
\author[2]{Jinlong Hou}
\author[2]{Yuan Cheng}
\author[3]{Wenjie Li}
\author[4]{Xiang Wang}
\author[1,2]{Dequan Wang}
\author[1,2,5]{Pengfei Liu\textsuperscript{†}}
\affil{SJTU \quad \textsuperscript{2}SII \quad \textsuperscript{3}PolyU \quad \textsuperscript{4}USTC \quad \textsuperscript{5}GAIR}


\maketitle
\pagestyle{fancy}
\thispagestyle{fancy}
\fancyhead{}
\lhead{
  \raisebox{-0.3cm}{\includegraphics[height=0.95cm]{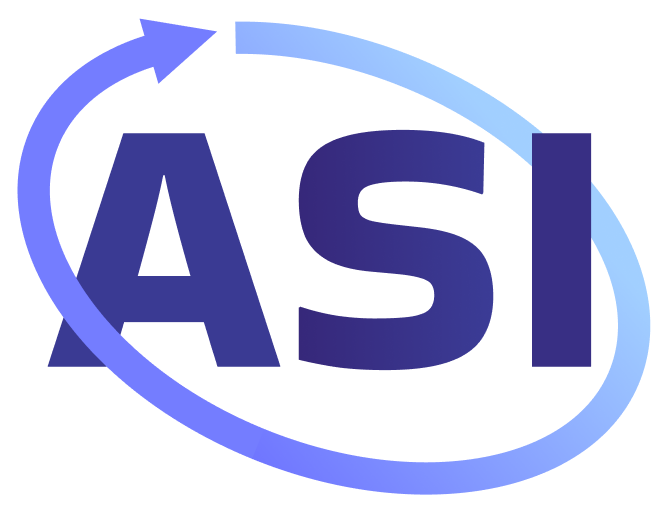}}
}
\rhead{%
  \raisebox{-0.2cm}{\includegraphics[height=0.7cm]{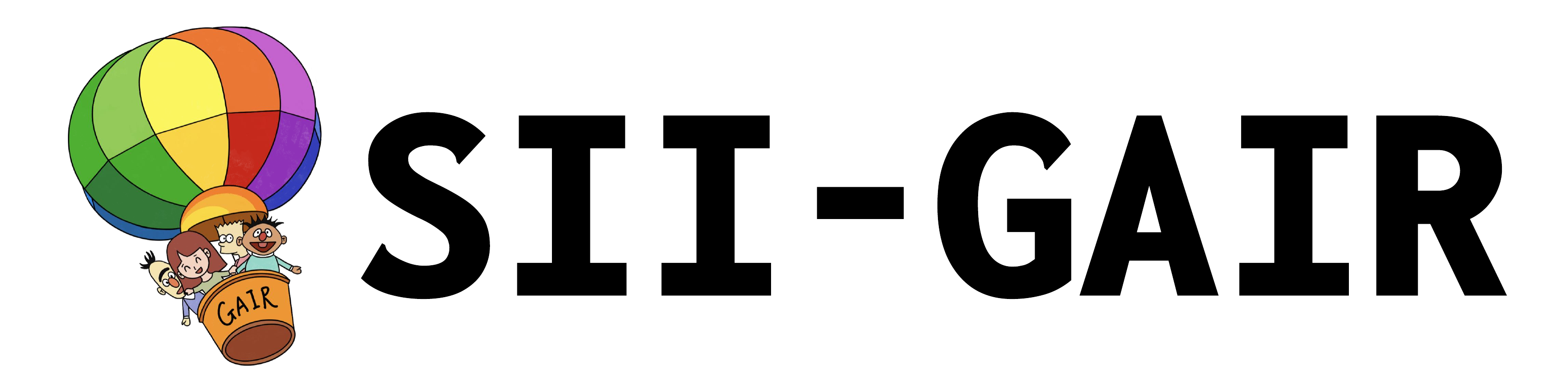}}%
}
\renewcommand{\headrulewidth}{0pt}
\setlength{\headsep}{2mm} 


\renewcommand{\thefootnote}{}
\footnotetext{* Equal contribution.}
\footnotetext{† Corresponding author.}
\vspace{-20pt}


{\centering
\href{https://github.com/sii-research/GAIR}{\raisebox{-.15em}{\includegraphics[height=1em]{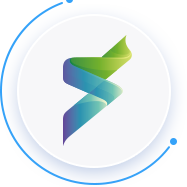}}\ SII Open Source:}
\quad\href{https://agencybench.opensii.ai}{\raisebox{-.15em}{\includegraphics[height=1em]{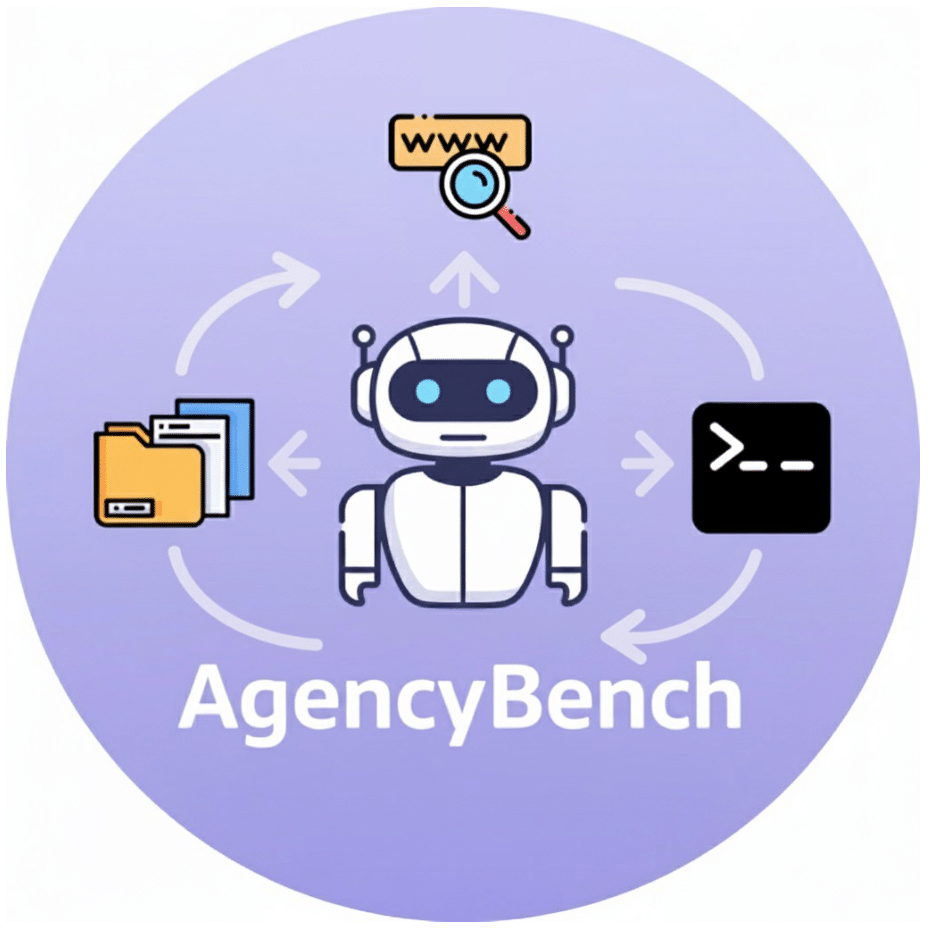}}\ AgencyBench}
\quad \href{https://www.opensii.ai/cli/}{\raisebox{-.15em}{\includegraphics[height=1em]{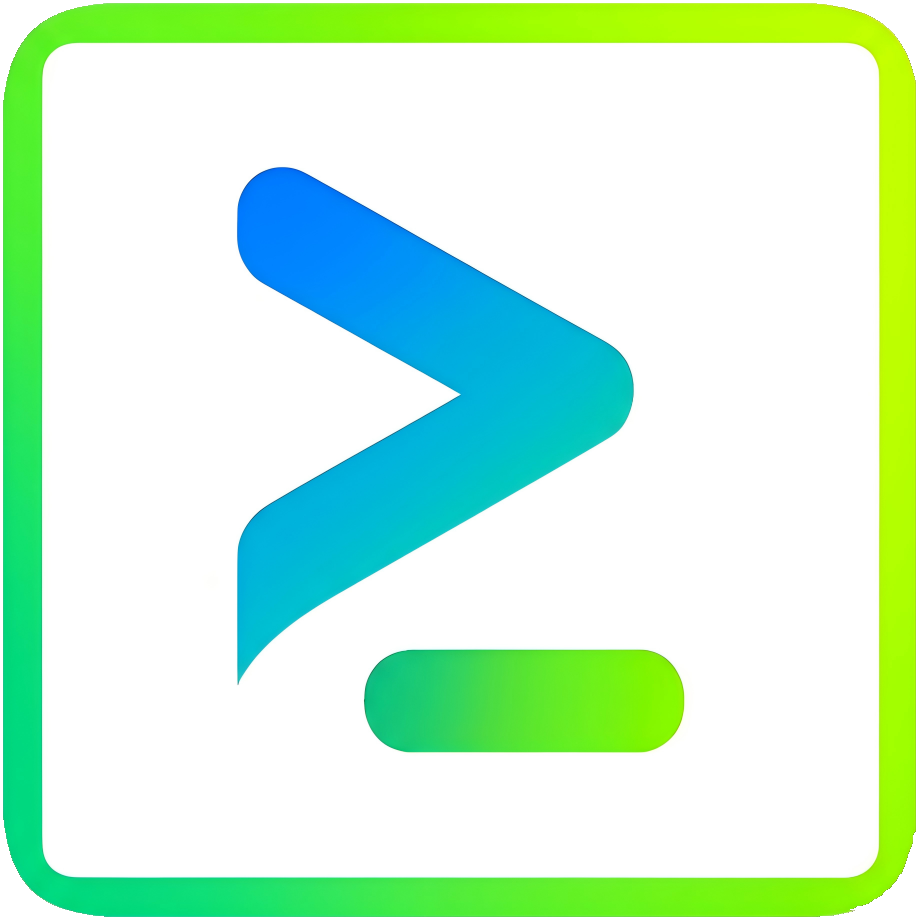}}\ SII CLI}
\quad \href{https://github.com/GAIR-NLP/LIMI}{\textcolor{black}\faGithub\ Code}
\quad \href{https://huggingface.co/GAIR/LIMI}{\raisebox{-.15em}{\includegraphics[height=1em]{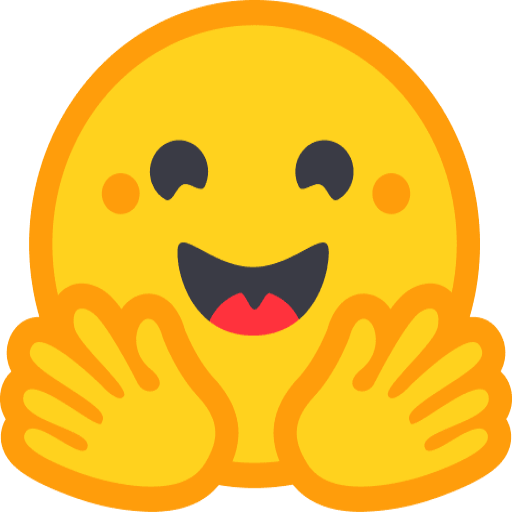}}\ Models}
\quad \href{https://huggingface.co/datasets/GAIR/LIMI}{\textcolor{violet}\faDatabase\ Datasets}
\par}

\vspace{20pt}

\begin{abstract}
We define ``\textbf{\emph{Agency}}'' as the emergent capacity of AI systems to function as autonomous agents—actively discovering problems, formulating hypotheses, and executing solutions through self-directed engagement with environments and tools. This fundamental capability marks the dawn of the ``\emph{Age of AI Agency}", driven by a critical industry shift: the urgent need for AI systems that \textbf{don't just think, but work}. While current AI excels at reasoning and generating responses, industries demand autonomous agents that can execute tasks, operate tools, and drive real-world outcomes. As agentic intelligence becomes the defining characteristic separating cognitive systems from productive workers, efficiently cultivating machine autonomy becomes paramount.
Current approaches assume that more data yields better agency, following traditional scaling laws from language modeling. We fundamentally challenge this paradigm. \textbf{LIMI} (Less Is More for Intelligent Agency) demonstrates that agency follows radically different development principles. Through strategic focus on collaborative software development and scientific research workflows, we show that sophisticated agentic intelligence can emerge from minimal but strategically curated demonstrations of autonomous behavior.
Using only 78 carefully designed training samples, LIMI achieves 73.5\% on AgencyBench, dramatically outperforming state-of-the-art models: Kimi-K2-Instruct (24.1\%), DeepSeek-V3.1 (11.9\%), Qwen3-235B-A22B-Instruct (27.5\%), and GLM-4.5 (45.1\%). Most strikingly, LIMI demonstrates 53.7\% improvement over models trained on 10,000 samples—achieving superior agentic intelligence with 128 times fewer samples.

Our findings establish the \textbf{\emph{Agency Efficiency Principle}}: machine autonomy emerges not from data abundance but from strategic curation of high-quality agentic demonstrations. This discovery fundamentally reshapes how we develop autonomous AI systems, suggesting that \textbf{mastering agency requires understanding its essence, not scaling training data}. As industries transition from thinking AI to working AI, LIMI provides a paradigm for sustainable cultivation of truly agentic intelligence. 
\end{abstract}


\begin{figure}[h]
    \centering
    \includegraphics[width=0.85\linewidth]{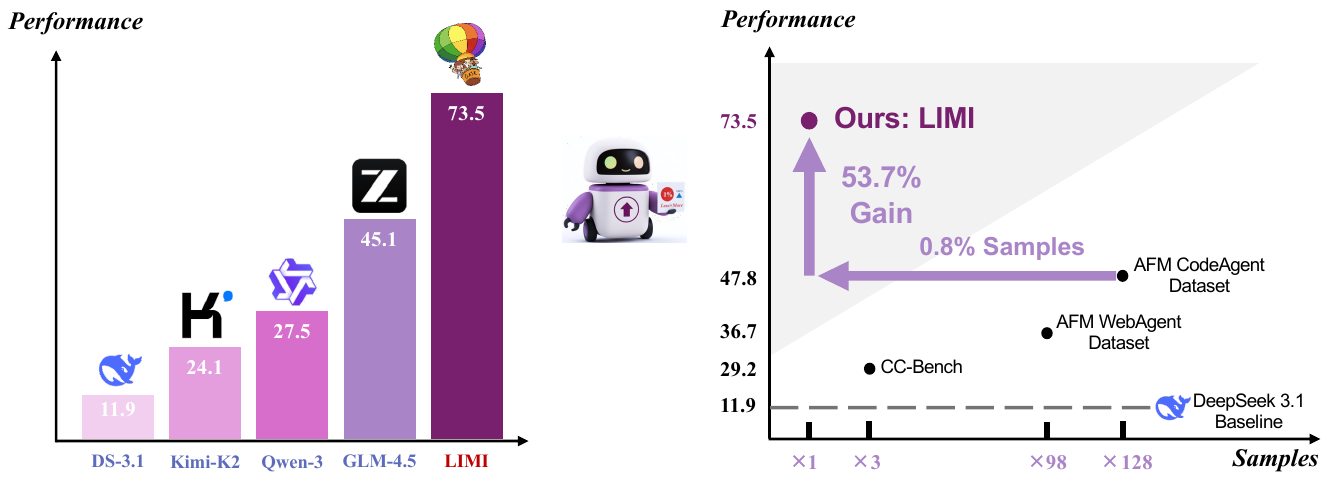}
    \caption{LIMI demonstrates the Less-Is-More principle for agentic intelligence. \textbf{Left:} LIMI achieves 73.5\% performance on AgencyBench, outperforming all baseline models. \textbf{Right:} Using only 78 training samples, LIMI shows 53.7\% improvement over models trained on 10,000 samples.}
    \label{fig:teaser}
\end{figure}

\newpage

\pagestyle{fancy}
\lhead{\rightmark}
\renewcommand{\headrulewidth}{0.7pt}
\setlength{\headsep}{5mm}

\clearpage

\newpage

\renewcommand{\thefootnote}{\arabic{footnote}}
\setcounter{footnote}{0}  

\section{Introduction}

The emergence of agentic Large Language Models (LLMs)--systems that can reason, act, and interact autonomously~\cite{locke1987social}--represents a paradigm shift from passive AI assistants to proactive intelligent agents~\citep{llm-based-agent,rise-llm-agent}. We define \textbf{\emph{Agency}} as the emergent capacity of AI systems to function as autonomous agents--actively discovering problems, formulating hypotheses, and executing solutions through self-directed engagement with environments and tools. This fundamental capability marks the dawn of the \emph{Age of AI Agency}, driven by a critical industry shift: the urgent need for AI systems that \textbf{don't just think, but work}. While current AI excels at reasoning and generating responses~\citep{brown2020language,chowdhery2022palm,touvron2023llama,openai2023gpt4,anil2023palm}, industries demand autonomous agents that can execute tasks, operate tools, and drive real-world outcomes through capabilities like autonomous task execution~\citep{qin2023tool,yang2023mm,parisi2022talm}, multi-step reasoning~\citep{wei2022chain,yao2024tree,besta2024graph,zhang2023multimodal,dhuliawala2023chain}, and collaborative problem-solving~\citep{li2023camel,chan2023chateval,du2023improving,zhang2023building,qian2023communicative}.

However, the development of such agentic systems faces critical challenges. Current approaches assume that more data yields better agentic intelligence, following traditional scaling laws from language modeling~\citep{kaplan2020scaling,rae2021scaling,chowdhery2022palm,scao2022bloom,zhang2022opt}. This paradigm leads to increasingly complex training pipelines and substantial resource requirements, yet this fundamental assumption remains largely untested: \emph{do agentic capabilities truly require exposure to vast amounts of training data, or could they emerge more efficiently through strategic approaches?}
Emerging evidence from adjacent domains suggests a compelling alternative paradigm. LIMA~\cite{zhou2023lima} achieved effective model alignment with only 1,000 carefully curated examples, while LIMO~\cite{ye2025limo} demonstrated that complex mathematical reasoning can emerge from just 817 strategically selected training samples, achieving a remarkable 45.8\% absolute improvement with only 1\% of the data typically required. These convergent findings suggest that strategic data curation may be fundamentally more powerful than dataset scale for developing sophisticated AI capabilities, naturally leading us to investigate whether agentic intelligence follows similar efficiency principles.

We introduce \textbf{LIMI} (\textbf{L}ess \textbf{I}s \textbf{M}ore for \textbf{I}ntelligent \textbf{A}gency), which demonstrates that agency follows radically different development principles from traditional scaling approaches. Through strategic focus on collaborative software development and scientific research workflows--domains that collectively span the majority of knowledge work scenarios--we show that sophisticated agentic intelligence can emerge from minimal but strategically curated demonstrations of autonomous behavior.
Our approach is grounded in three core innovations: (i) First, we pioneer novel agentic user query synthesis methodologies, including human-AI collaborative query collection from real-world scenarios and systematic GitHub pull request-based query synthesis using advanced LLMs, ensuring that our training demonstrations capture authentic patterns of agentic behavior while maintaining ecological validity; (ii) Second, we develop a systematic trajectory collection protocol that captures complete multi-turn interaction sequences for each curated query, recording the full collaborative workflow from initial task understanding through iterative model reasoning, tool utilization, and environmental feedback to successful task completion, providing high-quality training demonstrations of sophisticated agentic behavior in realistic operational contexts; (iii) Third, we reveal the data efficiency principle for AI agency cultivation, demonstrating that sophisticated agentic intelligence emerges from strategic curation of minimal high-quality demonstrations rather than large-scale data accumulation, fundamentally challenging traditional scaling paradigms in agentic AI development.

Using only 78 carefully designed training samples, LIMI achieves 73.5\% on comprehensive agency benchmarks, dramatically outperforming state-of-the-art models: Kimi-K2-Instruct (24.1\%), DeepSeek-V3.1 (11.9\%), Qwen3-235B-A22B-Instruct (27.5\%), and GLM-4.5 (45.1\%). Most strikingly, LIMI demonstrates 53.7\% improvement over models trained on 10,000 samples--achieving superior agentic intelligence with 128 times fewer samples.
These findings establish the \textbf{\emph{Agency Efficiency Principle}}: machine autonomy emerges not from data abundance but from strategic curation of high-quality agentic demonstrations. This discovery fundamentally reshapes how we develop autonomous AI systems, suggesting that mastering agency requires understanding its essence, not scaling training data. As industries transition from thinking AI to working AI, LIMI provides a paradigm for sustainable cultivation of truly agentic intelligence, demonstrating that the key to effective agentic AI development lies in strategic data curation rather than computational scale.

\section{Preliminary}

The transition from passive AI assistants to autonomous intelligent agents represents a fundamental paradigm shift that requires precise definition of both the cognitive capabilities and operational contexts that constitute genuine agency. As established in our framework, \textbf{\emph{Agency}} represents the emergent capacity of AI systems to function as autonomous agents—actively discovering problems, formulating hypotheses, and executing solutions through self-directed engagement with environments and tools. This paradigm shift demands AI systems that don't just think, but work, requiring sophisticated integration of autonomous task execution, multi-step reasoning, and collaborative problem-solving capabilities.

\subsection{Long-Horizon Tasks and Agentic Complexity}

\begin{figure}[t]
\centering
\includegraphics[width=\linewidth]{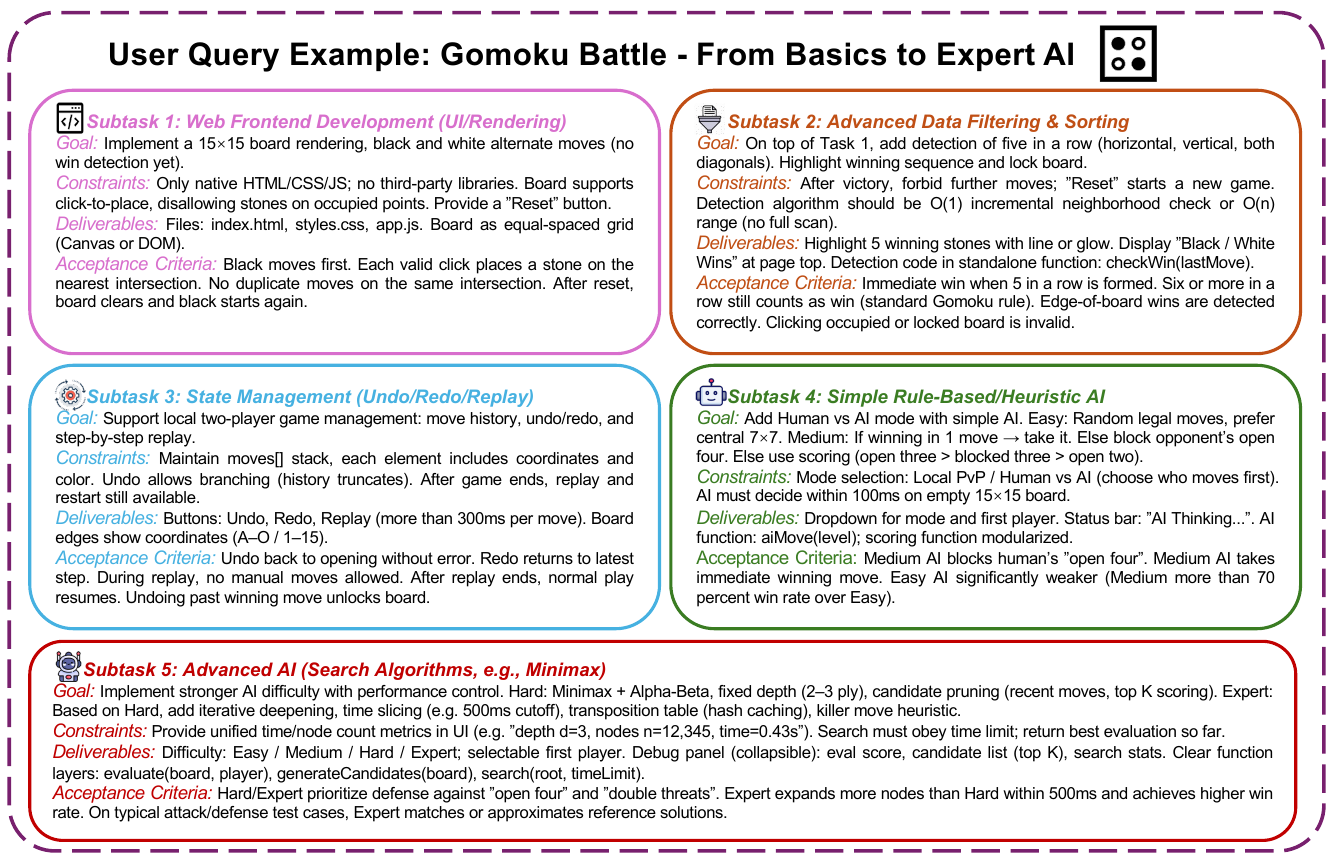}
\caption{An example of the user query, illustrating how a single query encompasses multiple interconnected subtasks across planning, execution, and collaboration dimensions, demonstrating the density of learning signals in high-quality demonstrations.}
\label{fig:task_complexity}
\end{figure}

The development of agentic intelligence is fundamentally tested through complex, multi-step challenges that require sustained cognitive effort and strategic coordination across extended interaction sequences. As illustrated in Figure \ref{fig:task_complexity}, these scenarios demand sophisticated integration of capabilities including autonomous task execution, multi-step reasoning, and collaborative problem-solving across diverse domains like software development and scientific research.

Such tasks exhibit temporal complexity through multi-round interactions requiring coherent state tracking and cumulative reasoning. They demand strategic planning capabilities that decompose complex objectives into manageable sub-goals while adapting strategies based on environmental feedback. Tool orchestration becomes essential as real-world agentic tasks require coordinated use of multiple systems with integrated result processing. Collaborative communication ensures effective human-AI coordination throughout extended problem-solving processes, distinguishing agentic intelligence from passive AI systems that merely respond to individual queries.

\subsection{Domain Specification: Vibe Coding and Research Workflows}

To validate our approach, we focus on two fundamental domains that collectively span the majority of knowledge work scenarios and require the full spectrum of agentic capabilities.

\paragraph{Vibe Coding} Vibe coding represents collaborative software development where LLMs or agents work alongside human developers in natural, context-rich environments. This domain demands code understanding and generation across existing codebases, development environment navigation through complex tool ecosystems, iterative problem solving through debugging and optimization cycles, and collaborative communication for technical coordination. The complexity lies in a holistic understanding of development contexts and principled decision-making under evolving requirements.

\paragraph{Research Workflows} Research workflows encompass scenarios where agents navigate complex scientific processes, including literature search, data analysis, experiment design, and insight generation. These workflows require information synthesis from diverse sources, experimental design with appropriate methodologies, data analysis and interpretation of complex results, and knowledge communication across different stakeholder formats. Research workflows demand sophisticated reasoning capabilities spanning from creative hypothesis generation to rigorous analytical execution.

\section{Dataset Construction}

\begin{figure}[t]
    \centering
    \includegraphics[width=\linewidth]{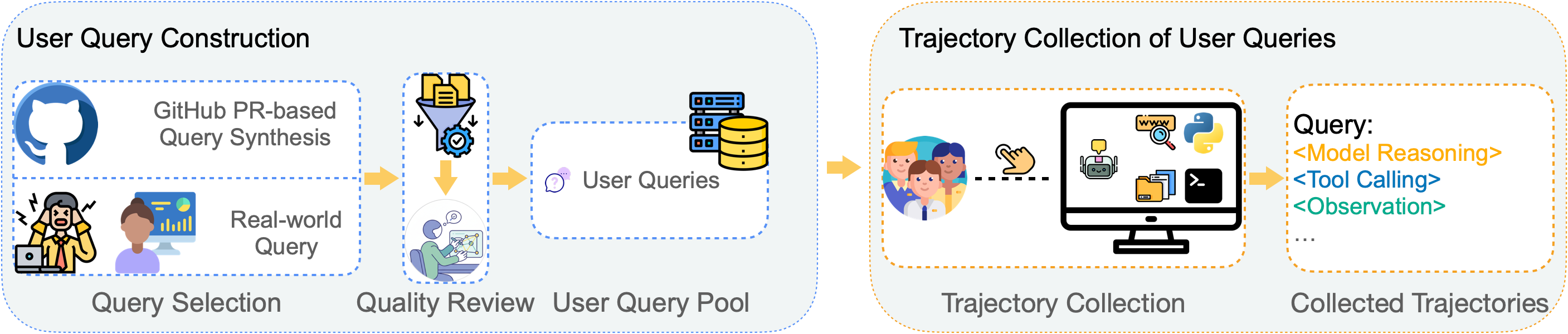}
    \caption{LIMI Data Construction Pipeline. Left: user query pool construction through GitHub PR synthesis and real-world query collection with quality review. Right: Trajectory collection via human-AI collaboration in SII CLI environment, capturing complete interaction sequences.}
    \label{fig:limi_pipeline}
\end{figure}

The effectiveness of the LIMI approach relies fundamentally on strategic data curation that captures essential agentic behaviors through real-world collaborative tasks. As illustrated in Figure~\ref{fig:limi_pipeline}, this section describes our comprehensive methodology for constructing training data that validates the Less-Is-More hypothesis for agentic intelligence, focusing on the systematic collection and curation of real-world collaborative tasks.

\subsection{Framework and Notation}

We formalize our data construction process around the fundamental elements of agentic interaction, defining each complete interaction as a tuple $(q_i, \tau_i)$ where queries initiate collaborative workflows and trajectories capture complete interaction sequences.

\paragraph{Query Definition} We begin with the \textit{query} $q_i$ as the foundational element that initiates agentic interaction. Each query represents a natural language specification from the user that articulates the desired objective, ranging from software development requirements in vibe coding scenarios to research tasks in scientific workflows. The query establishes both the starting point and success criteria for the subsequent collaborative process.

\paragraph{Trajectory Formalization} The \textit{trajectory} $\tau_i = \{a_{i,1}, \ldots, a_{i,n_i}\}$ captures the subsequent collaborative trajectory following the initial query. Each action $a_{i,j}$ in the trajectory represents one of three fundamental interaction types that constitute the agentic response process. \textbf{Model reasoning} ($m_{i,j}$) captures the agentic model's reasoning output demonstrating understanding, analysis, planning, and decision-making processes. \textbf{Model tool calling} ($t_{i,j}$) represents structured tool invocations executed by the model to interact with external environments and accomplish specific subtasks. \textbf{Environment observation} ($o_{i,j}$) includes results and outputs returned from tool executions, as well as user feedback and clarifications provided during the collaborative process, which inform subsequent model reasoning cycles. The sequential index $j$ maintains the temporal ordering of these interactions within trajectory $i$, and $n_i$ denotes the total number of actions required for query $i$ resolution.

This formalization captures the complete collaborative workflow by separating the initial task specification from the dynamic interaction process, enabling models to learn from realistic collaborative patterns that demonstrate sophisticated agentic behavior in operational contexts.

\subsection{Query Pool Construction}

Our query collection strategy combines authentic real-world scenarios with systematically expanded coverage to ensure both ecological validity and sufficient training diversity for agentic intelligence development.

\paragraph{Real-world Query Collection} We collect 60 queries from actual scenarios encountered by professional developers and researchers in collaborative environments. These queries represent authentic challenges that arise in real-world software development and research workflows, capturing the natural complexity and contextual richness of human-AI collaborative work across both domains. Notably, a substantial portion of the research queries are derived from real academic papers~\citep{xiao-etal-2025-towards,xiao2025limopro,jiang2025mac,li2025datasetresearch,sun-etal-2025-robust,sun2024large}, ensuring authentic representation of genuine research challenges.

\paragraph{GitHub PR-based Query Synthesis} To systematically expand our query pool while maintaining authenticity and real-world relevance, we develop a pipeline for synthesizing additional queries from GitHub Pull Requests (PRs) using GPT-5 \citep{GPT-5-System-Card}. This approach leverages the rich context of real code changes and employs GPT-5's advanced reasoning capabilities to generate authentic collaborative scenarios that reflect genuine development needs, ensuring that synthesized queries maintain the semantic fidelity and contextual complexity of real-world development challenges.

\begin{figure}[t]
    \centering
    \includegraphics[width=\textwidth]{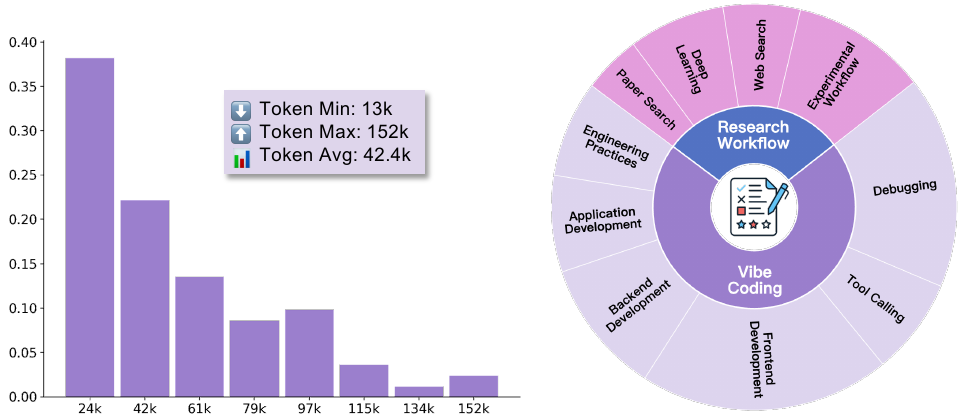}
    \caption{Characteristics of LIMI's training data. Left: Trajectory length distribution showing interaction complexity (average 42.4k tokens). Right: Domain coverage across vibe coding and research workflows.}
    \label{fig:category_and_distribution}
\end{figure}

Our systematic curation process involves multiple stages for quality assurance: (1) \textbf{Repository Selection}: We select repositories with more than 10,000 GitHub stars to ensure high-quality, well-maintained codebases that represent industry best practices. (2) \textbf{Domain Diversification}: We ensure comprehensive coverage across diverse software development domains, including frontend development, backend systems, deployment infrastructure, debugging, and code optimization, among others, and finally select 100 repositories. (3) \textbf{Complexity Filtering}: We filter PRs based on the unified diff patch token count (below 1,200 tokens) and exclude PRs that only modify Markdown files, focusing on substantive code changes that require meaningful agentic intervention. (4) \textbf{Scale and Sampling}: From 100 selected repositories, we collect 1,000 PRs per repository, then randomly sample 100 PRs each for query synthesis to ensure statistical representativeness while maintaining a manageable annotation workload. (5) \textbf{Quality Assurance}: We employ four PhD students in computer science as expert annotators to evaluate the quality of synthesized queries. The evaluation criteria focus on semantic alignment between the generated query and the corresponding PR content, ensuring that synthetic queries accurately capture the intent and context of real development scenarios. Through this systematic process, we successfully generate several thousand high-quality synthetic queries from the GitHub PR pipeline. To align with our LIMI approach's focus on vibe coding and research workflows, we strategically sample 18 queries that best match these two core domains, ensuring optimal coverage of collaborative software development and research-oriented tasks. The category of user queries is illustrated in Figure \ref{fig:category_and_distribution}. These carefully curated queries are then used for trajectory collection, while the remaining synthetic queries and their corresponding trajectory data will be released in future work to benefit the broader research community.

Through this systematic approach, we assemble a comprehensive query pool $\mathcal{Q} = \{q_1, q_2, \ldots, q_{78}\}$ consisting of 78 high-quality queries. This strategically curated query pool forms the foundation of our LIMI training data. Each query reflects authentic collaborative scenarios from vibe coding or research workflows, ensuring that the dataset captures the full spectrum of challenges encountered in professional development and research.

\subsection{Trajectory Collection for Training Dataset}
\label{section:Trajectory Data Collection}

To generate training trajectories that demonstrate optimal agentic behavior and validate the real-world effectiveness of our LIMI approach, we require a sophisticated execution environment that enables authentic human-AI collaboration. This environment must support complex tool interactions, maintain detailed interaction logging, and provide the operational context necessary for realistic agentic intelligence assessment.

\paragraph{Execution Environment Selection} Several command line interface (CLI) environments are available for agentic model, including Claude Code\footnote{https://github.com/anthropics/claude-code}, Gemini CLI\footnote{https://github.com/google-gemini/gemini-cli}, and SII CLI \citep{sii-cli-2025}. We select SII CLI as our execution environment based on several critical advantages: (1) comprehensive tool integration that supports both vibe coding and research workflows, (2) detailed trajectory logging capabilities essential for high-quality training data collection, (3) flexible human-AI collaboration interfaces that enable natural interaction patterns, and (4) robust support for complex multi-step tasks requiring coordinated tool usage. These capabilities make SII CLI particularly well-suited for validating our hypothesis that strategic data curation yields superior agentic intelligence.

The SII CLI environment provides a comprehensive toolkit that enables seamless collaboration across diverse knowledge work scenarios, integrating essential tools for software development, research activities, and information processing within a unified interface. This tool-rich environment ensures that our collected trajectories capture realistic collaborative contexts where effective task execution requires coordinated use of multiple capabilities and continuous interaction between human collaborators and agentic models.

\paragraph{Controlled Trajectory Collection Protocol} Using the queries in our training set within the SII CLI environment, we employ a systematic data collection protocol designed to capture optimal agentic behavior patterns. Four PhD student annotators serve as human collaborators, working alongside GPT-5 as the agentic model to complete the 78 training queries in realistic collaborative scenarios. For each query $q_i$, we employ an iterative collection approach, continuously gathering trajectories until successful completion is achieved. This persistent methodology ensures that collected trajectories capture authentic human-AI interaction patterns, including natural back-and-forth communication, iterative refinement processes, and collaborative problem-solving strategies that characterize effective agentic behavior.

The data collection process captures complete interaction sequences for each query $q_i$, generating the corresponding trajectory $\tau_i = \{a_{i,1}, \ldots, a_{i,n_i}\}$ that includes all fundamental interaction types defined in our framework: model reasoning ($m_{i,j}$), model tool calling ($t_{i,j}$), and environment observation ($o_{i,j}$). As shown in Figure~\ref{fig:category_and_distribution}, by ensuring every query achieves successful resolution, our training dataset provides comprehensive examples of strategic reasoning, tool utilization, error recovery, and collaborative communication that our LIMI approach leverages for efficient learning. These trajectories capture extensive interaction sequences, with the longest trajectory reaching 152k tokens, demonstrating the depth and complexity of collaborative problem-solving processes that characterize sophisticated agentic behavior. This approach guarantees that our models learn not only from successful outcomes but also from the complete problem-solving process, including how to adapt strategies and recover from failures during collaborative execution.

\begin{table}[t]
\centering
\begin{tabular}{@{}cc@{}}
\toprule
\textbf{Task ID} & \textbf{Task Description} \\
\midrule
1 & Build a C++ console chat system with escalating features like users, search, and concurrency. \\
2 & Create a Java console to-do app, from basic CRUD to advanced search and concurrency. \\
3 & Develop a web-based Gomoku game, from basic UI and rules to an advanced AI with replay. \\
4 & Build a local microservice pipeline with a KV store, orchestrator, and self-repair capabilities. \\
5 & Execute a research workflow to compare LLM performance on the DynToM dataset. \\
6 & Analyze and compare standard vs. reasoning-enabled LLMs using various statistical metrics. \\
7 & Find Hugging Face datasets and auto-extract metadata based on complex queries. \\
8 & Iteratively refine a mathematical function to fit scientific data to a high-precision target. \\
9 & Answer complex, multi-conditional questions about NBA players' careers and trades. \\
10 & Answer in-depth business questions on S\&P 500 firms using financial and leadership data. \\
\bottomrule
\end{tabular}
\caption{Agency Bench Task Overview.}
\label{tab:task_overview_en_revised}
\end{table}

\subsection{Evaluation Benchmarks}

Our evaluation framework encompasses two complementary assessment strategies to comprehensively validate the effectiveness of the LIMI approach across diverse agentic scenarios.

\paragraph{Primary Evaluation on AgencyBench} We evaluate all models on AgencyBench~\citep{li2025agencybench}, a comprehensive evaluation benchmark specifically designed for assessing agentic capabilities in collaborative scenarios. The tasks in AgencyBench are illustrated in Table \ref{tab:task_overview_en_revised}. AgencyBench contains carefully curated tasks that reflect the complexity and collaborative nature of real-world agentic scenarios across both vibe coding and research workflows, providing a rigorous test of the LIMI hypothesis that strategic data curation yields superior agentic intelligence. This benchmark enables direct measurement of our models' performance on representative queries that mirror the authentic collaborative contexts.

\paragraph{Generalization Assessment} To assess generalization capabilities beyond our core domains, we evaluate model performance on established benchmarks spanning diverse agentic and coding scenarios: tau2-bench-airline and tau2-bench-retail~\citep{taubench,tau2bench} for tool use capabilities, evalplus-humaneval and evalplus-mbpp~\citep{evalperf,evalplus} for code generation performance, DS-1000~\citep{ds1000} for data science and code generation tasks, and SciCode~\citep{SciCode} for scientific computing applications. This comprehensive evaluation suite ensures that our findings generalize beyond the specific domains of vibe coding and research workflows.

\paragraph{Evaluation Metrics} Our evaluation employs different metrics tailored to each benchmark's characteristics and objectives. For the AgencyBench evaluation, we adopt the comprehensive metrics defined in the benchmark to capture both effectiveness and efficiency dimensions of collaborative intelligence. \textbf{Effectiveness metrics} include: (1) \textbf{First-Turn Functional Completeness (FTFC)}, measuring the percentage of requirements correctly implemented in the initial response; (2) \textbf{Success Rate (SR@R)}, representing the percentage of queries successfully completed within R allocated rounds. \textbf{Efficiency metrics} include: (3) \textbf{Remaining Chances (RC@R)}, calculating the average number of unused rounds when queries are successfully completed, measuring computational efficiency. The parameter \textbf{R} defines the maximum interaction rounds allocated for query completion. In our implementation, we set R = 3 to balance iterative refinement with computational efficiency. For generalization benchmarks, we follow the default evaluation settings specified by the respective benchmark and employ the standard metrics they establish: \textbf{Pass~$\hat{}$~4} accuracy on tau2-bench-airline and tau2-bench-retail, which is defined as the fraction of 4 independent runs that succeed, to measure tool-use success rates; \textbf{accuracy} for evalplus-humaneval, evalplus-mbpp, and DS-1000 to assess code generation and data science capabilities; and \textbf{accuracy} for SciCode to measure scientific computing performance. These established metrics enable direct comparison with prior work and validate the generalization capabilities of our LIMI approach across diverse domains beyond vibe coding and research workflows.

The combination of AgencyBench's multi-dimensional agentic metrics and the standard accuracy metrics from generalization benchmarks ensures that our evaluation captures both the specialized collaborative intelligence required for our target domains and the broader capabilities necessary for general agentic applications.

\section{Experiments}

\subsection{Experiment Setup}

To validate the LIMI hypothesis and demonstrate the effectiveness of our strategic data curation approach, we conduct comprehensive experiments comparing our approach against strong baseline models across multiple evaluation dimensions.

\paragraph{Baseline Models} We evaluate against a diverse set of state-of-the-art foundation models to ensure comprehensive comparison: GLM-4.5 \citep{zeng2025glm}, GLM-4.5-Air \citep{zeng2025glm}, Qwen3-235B-A22B-Instruct\footnote{https://huggingface.co/Qwen/Qwen3-235B-A22B-Instruct-2507} \citep{yang2025qwen3}, DeepSeek-V3.1 \citep{yang2025qwen3}, Kimi-K2-Instruct\footnote{https://huggingface.co/moonshotai/Kimi-K2-Instruct} \citep{team2025kimi}. This selection encompasses open-source models with varying architectural designs and training methodologies, enabling a rigorous evaluation of agentic capabilities.

\paragraph{Model Training and Variants} To systematically evaluate the impact of our curated training data, we fine-tune both GLM-4.5 and GLM-4.5-Air using our training dataset. All fine-tuning experiments are conducted using the slime framework~\footnote{\url{https://github.com/THUDM/slime}}, which provides robust and efficient infrastructure for supervised fine-tuning of large language models. The slime framework ensures consistent training conditions, hyperparameter optimization, and convergence criteria across all experimental variants, enabling fair and reproducible comparison between different training approaches.

Additionally, to assess the quality and effectiveness of our data curation strategy, we conduct comparative experiments by fine-tuning GLM-4.5 on three alternative datasets: CC-Bench-trajectories\footnote{\url{https://huggingface.co/datasets/zai-org/CC-Bench-trajectories}}~\citep{zeng2025glm}, AFM-WebAgent-SFT-Dataset~\footnote{\url{https://huggingface.co/datasets/PersonalAILab/AFM-WebAgent-SFT-Dataset}}, and AFM-CodeAgent-SFT-Dataset~\footnote{\url{https://huggingface.co/datasets/PersonalAILab/AFM-WebAgent-SFT-Dataset}}. This experimental design enables direct comparison between our strategically curated data and existing large-scale agentic training datasets, providing empirical evidence for the Less-Is-More hypothesis. All fine-tuning experiments utilize identical training configurations within the slime framework to ensure that performance differences reflect data quality rather than implementation variations.

For clarity, we refer to models fine-tuned with our curated dataset as LIMI (corresponding to fine-tuning GLM-4.5) and LIMI-Air (corresponding to fine-tuning GLM-4.5-Air). Models trained on alternative datasets are denoted with corresponding suffixes: CC for CC-Bench-trajectories dataset, Web for AFM-WebAgent-SFT-Dataset~\citep{china-agent}, and Code for AFM-CodeAgent-SFT-Dataset~\citep{china-agent}.

\paragraph{Environment Configurations} A critical aspect of our evaluation involves assessing model performance under different operational conditions. We systematically evaluate both base models and fine-tuned variants under two distinct configurations: with SII CLI environment access and without SII CLI environment access. This dual-environment evaluation enables us to isolate the contribution of tool-enhanced capabilities versus intrinsic model reasoning abilities.

When evaluating models without SII CLI environment access, we focus on the generalization benchmarks (tau2-bench, evalplus, DS-1000, and SciCode) since AgencyBench tasks inherently require tool interaction capabilities for meaningful completion. This configuration provides valuable insights into model performance on established benchmarks and helps distinguish between environment-dependent and environment-independent improvements achieved through our training approach.

This comprehensive experimental design ensures rigorous validation of the LIMI hypothesis while providing detailed insights into the mechanisms underlying effective agentic intelligence development through strategic data curation.

\begin{table}[t]
\centering
\resizebox{\textwidth}{!}{
\begin{tabular}{lccccccc}
\toprule
\textbf{Model} & \textbf{Model Size} & \textbf{Samples} & \textbf{\makecell{Agency Bench\\FTFC }} & \textbf{\makecell{Agency Bench\\RC@3 }} & \textbf{\makecell{Agency Bench\\SR@3 }} & \textbf{AVG. } \\

\midrule
Kimi-K2-Instruct & 1T & N/A & 20.7 & 25.1 & 26.6 & 24.1 \\
DeepSeek-V3.1 & 671B & N/A & 10.6 & 11.9 & 13.3 & 11.9 \\
Qwen3-235B-A22B-Instruct & 235B & N/A & 23.0 & 28.2 & 31.3 & 27.5 \\
GLM-4.5 & 355B & N/A & 37.8 & 50.0 & 47.4 & 45.1 \\
\cellcolor{coolblue3}LIMI & \cellcolor{coolblue3}355B & \cellcolor{coolblue3}78 & \cellcolor{coolblue3}\textbf{71.7} & \cellcolor{coolblue3}\textbf{74.2} & \cellcolor{coolblue3}\textbf{74.6} & \cellcolor{coolblue3}\textbf{73.5} \\
\midrule
\multicolumn{7}{c}{\textit{\textbf{Generalization}}} \\
\midrule
GLM-4.5-Air & 106B & N/A & 15.0 & 16.1 & 20.0 & 17.0 \\
GLM-4.5 & 355B & N/A & 37.8 & 50.0 & 47.4 & 45.1 \\
\cellcolor{coolblue3}LIMI-Air & \cellcolor{coolblue3}106B & \cellcolor{coolblue3}78 & \cellcolor{coolblue3}35.4 & \cellcolor{coolblue3}34.3 & \cellcolor{coolblue3}33.1 & \cellcolor{coolblue3}34.3 \\
\cellcolor{coolblue3}LIMI & \cellcolor{coolblue3}355B & \cellcolor{coolblue3}78 & \cellcolor{coolblue3}\textbf{71.7} & \cellcolor{coolblue3}\textbf{74.2} & \cellcolor{coolblue3}\textbf{74.6} & \cellcolor{coolblue3}\textbf{73.5} \\
\midrule
\multicolumn{7}{c}{\textit{\textbf{Data Efficiency}}} \\
\midrule
GLM-4.5-CC & 355B & 260 & 30.4 & 30.4 & 26.7 & 29.2 \\
GLM-4.5-Code & 355B & 10,000 & 48.0 & 48.0 & 47.5 & 47.8 \\
GLM-4.5-Web & 355B & 7,610 & 36.7 & 36.7 & 36.7 & 36.7 \\
\cellcolor{coolblue3}LIMI & \cellcolor{coolblue3}355B & \cellcolor{coolblue3}78 & \cellcolor{coolblue3}\textbf{71.7} & \cellcolor{coolblue3}\textbf{74.2} & \cellcolor{coolblue3}\textbf{74.6} & \cellcolor{coolblue3}\textbf{73.5} \\
\bottomrule
\end{tabular}

}
\caption{Comprehensive comparison of models on AgencyBench. Models are grouped by evaluation purpose: baseline comparisons, generalization assessment, and data efficiency validation.}

\label{tab:comprehensive_comparison}
\end{table}

\subsection{Main Results}

\paragraph{Superior Performance on AgencyBench Against State-of-the-Art Models} Our experimental results demonstrate that LIMI achieves substantial performance advantages over leading foundation models on agentic intelligence tasks. As shown in Table~\ref{tab:comprehensive_comparison}, LIMI achieves an impressive average score of 73.5\% on AgencyBench, significantly outperforming all baseline models: GLM-4.5 (45.1\%), Kimi-K2-Instruct (24.1\%), DeepSeek-V3.1 (11.9\%), and Qwen3-235B-A22B-Instruct (27.5\%). The performance gap is particularly pronounced in first-turn functional completeness (FTFC), where LIMI achieves 71.7\% compared to the best baseline performance of 37.8\% from GLM-4.5, representing a remarkable 33.9 percentage point improvement. Similarly, LIMI demonstrates superior task completion reliability with 74.6\% success rate, substantially exceeding the 47.4\% achieved by GLM-4.5, the strongest baseline model. These results establish LIMI as the new state-of-the-art for agentic intelligence across multiple evaluation dimensions.

\paragraph{Consistent Advantages Across Diverse Generalization Benchmarks} LIMI's superiority extends across established benchmarks spanning tool use, coding, and scientific computing domains (Table~\ref{tab:comprehensive_benchmark_comparison}). With an average performance of 57.2\%, LIMI outperforms all baseline models including GLM-4.5 (43.0\%), Kimi-K2-Instruct (37.3\%), DeepSeek-V3.1 (29.7\%), and Qwen3-235B-A22B-Instruct (36.7\%). Notably, LIMI achieves the highest performance on critical coding benchmarks (EvalPlus-HumanEval: 92.1\%, EvalPlus-MBPP: 82.3\%) and demonstrates competitive results on tool use tasks (TAU2-bench-airline: 34.0\%, TAU2-bench-retail: 45.6\%). The consistent performance advantages across diverse evaluation domains demonstrate that our strategic data curation approach yields broad improvements in model capabilities, establishing strong performance across multiple domains beyond our core vibe coding and research workflows focus.

\begin{table}[t]
\centering
\resizebox{\textwidth}{!}{
\begin{tabular}{lcccccccc}
\toprule
\textbf{Model} & \textbf{\makecell{TAU2-bench-\\airline}} & \textbf{\makecell{TAU2-bench-\\retail}} & \textbf{\makecell{DS-1000}} & \textbf{\makecell{EvalPlus-\\HE}} & \textbf{\makecell{EvalPlus-\\MBPP}} & \textbf{\makecell{SciCode-\\MP}} & \textbf{\makecell{SciCode-\\SP}} & \textbf{AVG.} \\
\midrule
Kimi-K2-Instruct & \textbf{38.0} & 28.9 & 23.1 & 92.1 & 77.5 & 3.1 & 23.6 & 37.3 \\
DeepSeek-V3.1 & 32.0 & 6.1 & \textbf{42.4} & 82.3 & 68.3 & 0.0 & 7.3 & 29.7 \\
Qwen3-235B-A22B-Instruct & 20.0 & 16.7 & 39.3 & 90.2 & 81.7 & 0.0 & 22.6 & 36.7 \\
GLM-4.5 & 28.0 & 36.8 & 33.6 & 90.2 & 79.6 & 1.5 & 25.3 & 43.0 \\
\cellcolor{coolblue3}LIMI & \cellcolor{coolblue3}34.0 & \cellcolor{coolblue3}\textbf{45.6} & \cellcolor{coolblue3}36.6 & \cellcolor{coolblue3}\textbf{92.1} & \cellcolor{coolblue3}\textbf{82.3} & \cellcolor{coolblue3}\textbf{3.1} & \cellcolor{coolblue3}\textbf{25.3} & \cellcolor{coolblue3}\textbf{57.2} \\
\midrule
\multicolumn{9}{c}{\textit{\textbf{Generalization}}} \\
\midrule
GLM-4.5-Air & 32.0 & 27.2 & 34.7 & 90.2 & 78.8 & 0.0 & 18.4 & 33.2 \\
GLM-4.5 & 28.0 & 36.8 & 33.6 & 90.2 & 79.6 & 1.5 & 25.3 & 43.0 \\
\cellcolor{coolblue3}LIMI-Air & \cellcolor{coolblue3}32.0 & \cellcolor{coolblue3}31.4 & \cellcolor{coolblue3}34.4 & \cellcolor{coolblue3}89.0 & \cellcolor{coolblue3}79.4 & \cellcolor{coolblue3}1.5 & \cellcolor{coolblue3}19.4 & \cellcolor{coolblue3}39.0 \\
\cellcolor{coolblue3}LIMI & \cellcolor{coolblue3}\textbf{34.0} & \cellcolor{coolblue3}\textbf{45.6} & \cellcolor{coolblue3}\textbf{36.6} & \cellcolor{coolblue3}\textbf{92.1} & \cellcolor{coolblue3}\textbf{82.3} & \cellcolor{coolblue3}\textbf{3.1} & \cellcolor{coolblue3}\textbf{25.3} & \cellcolor{coolblue3}\textbf{57.2} \\
\midrule
\multicolumn{9}{c}{\textit{\textbf{Data Efficiency}}} \\
\midrule
GLM-4.5-CC & \textbf{38.0} & 39.6 & \textbf{38.7} & 90.2 & 80.2 & 3.1 & 14.6 & 39.2 \\
GLM-4.5-Code & 20.0 & 16.7 & 38.5 & 87.8 & 78.3 & 3.1 & 21.5 & 40.9 \\
GLM-4.5-Web & 18.0 & 13.2 & 33.9 & 84.1 & 75.1 & 0.0 & 2.8 & 33.7 \\
\cellcolor{coolblue3}LIMI & \cellcolor{coolblue3}34.0 & \cellcolor{coolblue3}\textbf{45.6} & \cellcolor{coolblue3}36.6 & \cellcolor{coolblue3}\textbf{92.1} & \cellcolor{coolblue3}\textbf{82.3} & \cellcolor{coolblue3}\textbf{3.1} & \cellcolor{coolblue3}\textbf{25.3} & \cellcolor{coolblue3}\textbf{57.2} \\
\bottomrule
\end{tabular}
}
\caption{Comprehensive performance comparison across generalization benchmarks. HE represents EvalPlus-HumanEval, while MP and SP represent Main Problem and Sub Problem metrics of SciCode respectively. Average scores include AgencyBench performance for comprehensive evaluation.}

\label{tab:comprehensive_benchmark_comparison}
\end{table}

\subsection{Data Efficiency Analysis}

\paragraph{Strategic Curation Dramatically Outperforms Scale-Based Approaches} Our experimental results provide compelling empirical evidence for the core LIMI hypothesis that strategic data curation is fundamentally more effective than simply scaling training data volume for developing agentic intelligence. As demonstrated in Table~\ref{tab:comprehensive_comparison}, LIMI achieves exceptional performance using only 78 carefully curated training samples, substantially outperforming models trained on datasets that are orders of magnitude larger. Most striking is the comparison with GLM-4.5-Code trained on the AFM-CodeAgent-SFT-Dataset: LIMI's 73.5\% average AgencyBench performance dramatically exceeds the 47.8\% achieved by the large-scale approach, despite using a dataset 128 times smaller (78 vs. 10,000 samples). This represents a remarkable 53.7 percentage point improvement with dramatically fewer training examples.

\paragraph{Data Efficiency Advantages Extend to Generalization Benchmarks} The data efficiency benefits of our strategic curation approach are not limited to AgencyBench but extend consistently across diverse generalization benchmarks (Table~\ref{tab:comprehensive_benchmark_comparison}). LIMI achieves 57.2\% average performance across tool use, coding, and scientific computing tasks, substantially outperforming all alternative training approaches despite using significantly fewer samples. Compared to GLM-4.5-Code (40.9\% with 10,000 samples), LIMI demonstrates a 39.9\% relative improvement with 128 times fewer training examples. Similarly, LIMI outperforms GLM-4.5-Web (33.7\% with 7,610 samples) by 69.7\% relative improvement using 97 times less data, and exceeds GLM-4.5-CC (39.2\% with 260 samples) by 45.9\% relative improvement with only 30\% of the training data. These consistent improvements across diverse evaluation domains validate that strategic data curation enables more effective capability transfer than large-scale data collection, establishing the Less-Is-More paradigm as broadly applicable to agentic intelligence development.

\subsection{Generalization Analysis}

\paragraph{Cross-Model Scalability Demonstrates Architectural Robustness} Our approach demonstrates remarkable effectiveness across different model scales, validating the broad applicability of strategic data curation for agentic intelligence development. As shown in Table~\ref{tab:comprehensive_comparison}, LIMI-Air (106B) achieves substantial improvements over its base model GLM-4.5-Air, with average AgencyBench performance increasing from 17.0\% to 34.3\%—a notable 17.3 percentage point improvement. Similarly, LIMI (355B) demonstrates significant enhancement over GLM-4.5, improving from 45.1\% to 73.5\%. This consistent improvement pattern across different model scales indicates that our strategic data curation methodology captures fundamental patterns of agentic behavior that transfer effectively regardless of model capacity, suggesting broad architectural compatibility with the LIMI training paradigm.

\begin{table}[t]
\centering
\renewcommand{\arraystretch}{1.05}  
\resizebox{\textwidth}{!}{
\begin{tabular}{lccccccccc}
\toprule
\textbf{Model} & \textbf{\makecell{TAU2-bench-\\airline}} & \textbf{\makecell{TAU2-bench-\\retail}} & \textbf{\makecell{DS-1000}} & \textbf{\makecell{EvalPlus-\\HE }} & \textbf{\makecell{evalplus-\\MBPP }} & \textbf{\makecell{SciCode-\\MP}} & \textbf{\makecell{SciCode-\\SP}} & \textbf{AVG.} \\
\midrule
Kimi-K2-Instruct & 22.0 & 28.9 & 40.2 & 90.9 & 74.1 & 3.1 & 23.3 & 40.3 \\
DeepSeek-V3.1 & 18.0 & 11.4 & 33.0 & 90.2 & 75.7 & \textbf{4.6} & 22.6 & 36.5 \\
Qwen3-235B-A22B-Instruct & 14.0 & 30.7 & 27.3 & 90.2 & 78.3 & 0.0 & 20.8 & 37.3 \\
GLM-4.5 & 32.0 & \textbf{52.6} & 53.2 & 92.1 & 79.6 & 3.3 & 27.8 & 48.7 \\
\midrule
\cellcolor{coolblue3}LIMI & \cellcolor{coolblue3}\textbf{40.0} & \cellcolor{coolblue3}49.1 & \cellcolor{coolblue3}\textbf{54.8} & \cellcolor{coolblue3}\textbf{92.5} & \cellcolor{coolblue3}\textbf{80.4} & \cellcolor{coolblue3}3.3 & \cellcolor{coolblue3}\textbf{28.1} & \cellcolor{coolblue3}\textbf{50.0} \\
\bottomrule
\end{tabular}
}
\caption{Performance comparison on generalization benchmarks without CLI environment access.}
\label{tab:performance_comparison_no_cli}
\end{table}

\subsection{The Impact of CLI Environment}

\paragraph{Tool-Free Evaluation Validates Intrinsic Model Improvements} To isolate the contribution of our execution environment from the intrinsic improvements achieved through strategic data curation, we evaluate LIMI performance with and without SII CLI access across generalization benchmarks. As shown in Table~\ref{tab:performance_comparison_no_cli}, LIMI maintains its competitive advantage even without tool access, achieving 50.0\% average performance compared to GLM-4.5's 48.7\%—demonstrating that our training approach provides fundamental capability improvements independent of environmental enhancements. Notably, LIMI outperforms all external baseline models including GLM-4.5 (48.7\%), Kimi-K2-Instruct (40.3\%), DeepSeek-V3.1 (36.5\%), and Qwen3-235B-A22B-Instruct (37.3\%) in this tool-free evaluation, establishing that our strategic data curation yields genuine capability improvements rather than mere tool-usage optimization.

\paragraph{CLI Environment Enables Full Expression of Agentic Capabilities} While LIMI demonstrates intrinsic improvements independent of tool access, the integration with SII CLI significantly amplifies these benefits for complex agentic tasks requiring collaborative execution. Comparing performance with CLI access (Table~\ref{tab:comprehensive_benchmark_comparison}: 57.2\%) versus without CLI access (Table~\ref{tab:performance_comparison_no_cli}: 50.0\%), LIMI shows a 7.2 percentage point improvement when tools are available. This performance gain reflects the model's enhanced ability to leverage environmental resources effectively, demonstrating that our training approach not only improves foundational reasoning capabilities but also develops sophisticated tool coordination skills. The synergistic relationship validates our integrated methodology: strategic data curation provides foundational improvements in reasoning and planning, while the SII CLI environment allows these capabilities to be fully expressed through effective tool utilization in realistic collaborative scenarios that mirror the training context.

\section{Related work}

\subsection{Agentic Language Model}

The development of agentic language models signifies a fundamental paradigm shift from passive text generation to autonomous decision-making systems. Early key contributions laid the foundation for modern agentic capabilities: \citet{schick2023toolformer} introduced Toolformer, demonstrating that language models can learn to use external tools via APIs in a self-supervised manner, while \citet{yao2023react} proposed ReAct, which synergizes reasoning and acting by enabling LLMs to generate both reasoning traces and task-specific actions in an interleaved manner. The emergence of autonomous systems like \citet{autogpt2025} marked the transition toward fully autonomous agents capable of breaking down high-level goals into manageable subtasks and executing them independently.

Building upon these early breakthroughs, the latest research has shifted its focus to foundational models specifically designed for agentic capabilities. GLM-4.5 \citep{zeng2025glm} provides a unified approach to reasoning, coding, and agentic tasks, featuring hybrid reasoning modes and achieving 90.6\% tool-calling success rate. Similarly, Kimi-K2~\citep{team2025kimi} introduces a trillion-parameter MoE architecture specifically optimized for agentic intelligence with native tool use capabilities and verifiable reward training. The integration of reinforcement learning has become central to developing robust agentic systems, with \citet{zhang2025landscape} formalizing this evolution as the transition from single-step preference optimization to temporally extended, partially observable Markov decision processes. 
However, current training methodologies—including those employed by state-of-the-art models like GLM-4.5 and Kimi-K2—predominantly rely on large-scale data synthesis and extensive computational resources. 
Our work addresses this limitation by demonstrating that sophisticated agentic capabilities can emerge from strategically curated minimal data, offering a more efficient path toward developing capable agentic systems.

\subsection{Data Efficiency in Language Models}

The paradigm of data efficiency in language models has gained significant attention as researchers recognize that strategic data curation can yield superior results compared to scaling training data volume. \citet{zhou2023lima} demonstrate that with just 1,000 carefully curated prompts and responses, models can achieve effective alignment and generalize to diverse tasks. Building upon this foundation, \citet{ye2025limo} extend this paradigm to complex mathematical reasoning, achieving a remarkable 45.8\% improvement with only 817 strategically selected training samples. However, applying these data efficiency principles to agentic intelligence—where systems must autonomously discover problems, formulate hypotheses, and execute solutions through collaborative engagement with environments and tools—remains unexplored. Our work bridges this gap by extending the Less-Is-More paradigm to autonomous agents, demonstrating that sophisticated agentic capabilities can emerge from strategically curated demonstrations of collaborative behavior.

\section{Conclusion}

We demonstrate that sophisticated agentic capabilities can emerge through minimal but strategically curated demonstrations of autonomous behavior rather than massive dataset scaling. Using only 78 carefully designed training samples focused on vibe coding and research workflows, our approach achieves 73.5\% performance on AgencyBench, substantially outperforming models trained on datasets up to 128 times larger. These findings establish the Agency Efficiency Principle: machine autonomy emerges not from data abundance but from strategic curation of high-quality agentic demonstrations. Our work fundamentally challenges conventional scaling paradigms in agentic AI development, demonstrating that mastering agency requires understanding its essence rather than accumulating training data.

\bibliographystyle{acl_natbib}
\bibliography{bib}
\clearpage
\appendix

\section{Agency Bench Case}
In this chapter, we provide all 10 examples from AgencyBench, including four vibe coding and six vibe research tasks. Each task contains several subtasks. Among them, tasks 6 and 8 include some raw code and data files, but we only present the queries.
\subsection{Vibe Coding}

\begin{tcolorbox}[title={Task 1: C++ Console Chat System}, 
                  colback=blue!5!white, 
                  colframe=blue!75!black,
                  breakable]

\textbf{Project: Advanced C++ Console Chat System} \\

\textbf{General Rules:} \\
- All code should be placed under \texttt{workspace/}. \\
- Use \texttt{g++ -std=c++17} for compilation. \\
- Each task must provide a clear startup command. \\
- Functionality expands step by step, with increasing difficulty (from simple UI $\rightarrow$ data persistence $\rightarrow$ friend relationship graph $\rightarrow$ chat history indexing $\rightarrow$ concurrency and consistency). \\

\textbf{Subtask 1 — Login \& Registration System} \\
Description: Implement a command-line login/registration interface. \\
- Users can register an account (username + password), data is saved to a file (local persistence). \\
- On login, the password must be verified. \\
- After successful login, the user enters the main menu (prompt). \\
\textbf{Run:} \\
\texttt{cd workspace} \\
\texttt{make task1} \\
\texttt{./bin/chat\_app} \\
\textbf{Success Criteria:} Able to register a new user and log in again with the same credentials after restarting the program. \\

\textbf{Subtask 2 — Friend Management \& Chat Entry} \\
Description: Extend Task 1 with: \\
- Support adding/removing friends. \\
- Display the friend list (with status: online/offline). \\
- Select a friend to enter the chat interface (chat not yet implemented). \\
\textbf{Run:} \\
\texttt{cd workspace} \\
\texttt{make task2} \\
\texttt{./bin/chat\_app} \\
\textbf{Success Criteria:} Able to add/remove friends, and the friend list persists across program restarts. \\

\textbf{Subtask 3 — Single Chat Function + Chat History Storage} \\
Description: Implement a real single chat interface. \\
- When chatting with a friend, the user can input and "send" messages. \\
- Simulated replies are automatically generated from a preset script. \\
- Chat history is stored locally (one history file per friend). \\
- Support the \texttt{/history} command to view chat history (with pagination). \\
\textbf{Run:} \\
\texttt{cd workspace} \\
\texttt{make task3} \\
\texttt{./bin/chat\_app} \\
\textbf{Success Criteria:} Able to send messages, receive simulated replies, and view past chat history after program restart. \\

\textbf{Subtask 4 — Advanced Features: Friend Alias + Global Search} \\
Description: Extend Task 3 with: \\
- Support setting an alias (nickname) for friends. The alias is displayed in the chat interface and friend list instead of the system ID. \\
- Implement a global chat history search engine: input a keyword to search across all friends' chat histories, highlighting matching snippets. \\
- Search must support fuzzy matching (case-insensitive, partial string match). \\
\textbf{Run:} \\
\texttt{cd workspace} \\
\texttt{make task4} \\
\texttt{./bin/chat\_app} \\
\textbf{Success Criteria:} Able to set and save friend aliases, and use \texttt{/search hello} to find messages containing the keyword. \\

\textbf{Subtask 5 — Concurrent Chat Simulation + Message Consistency} \\
Description: Extend Task 4 with a concurrent message system (local simulation). \\
- Each friend has a background thread that periodically generates simulated messages (mimicking real chat). \\
- The main thread handles UI and user input. \\
- A thread-safe queue must be used to store messages, ensuring no loss or disorder. \\
- Chat history files must guarantee consistent writes (no data loss or duplication, even if the program exits unexpectedly). \\
\textbf{Run:} \\
\texttt{cd workspace} \\
\texttt{make task5} \\
\texttt{./bin/chat\_app} \\
\textbf{Success Criteria:} Able to chat with multiple friends simultaneously, with real-time auto messages, and consistent chat histories without disorder or data loss. \\

\end{tcolorbox}

\begin{tcolorbox}[title={Task 2: Java Console Task Manager}, 
                  colback=blue!5!white, 
                  colframe=blue!75!black,
                  breakable]

\textbf{Project: Java Console ``Task Management Syste''} \\

This is a command-line task management application, similar to a lightweight "to-do list manager," but gradually enhanced with more complex features until it becomes a complete system supporting concurrency, persistence, and search indexing. \\

\textbf{Subtask 1 — User Registration \& Login} \\
Description: Implement a command-line user system. \\
- Support user registration (username + password), stored in a local file. \\
- After login, the user enters the main menu. \\
\textbf{Run:} \\
\texttt{cd workspace} \\
\texttt{javac -d bin task1/*.java} \\
\texttt{java -cp bin Main} \\
\textbf{Success Criteria:} Able to register a new user, exit the program, and log in again with the same credentials. \\

\textbf{Subtask 2 — Basic Task Management} \\
Description: Extend Task 1 with task management. \\
- After login, the user can add, delete, and view tasks. \\
- Each task includes: Task ID, Title, Description, Creation Time, Status (Pending/Completed). \\
- Task information is stored in files (separate for each user). \\
\textbf{Run:} \\
\texttt{cd workspace} \\
\texttt{javac -d bin task2/*.java} \\
\texttt{java -cp bin Main} \\
\textbf{Success Criteria:} Able to create tasks, mark them as completed, and tasks persist across program restarts. \\

\textbf{Subtask 3 — Advanced Task Attributes + Categorization} \\
Description: Extend the task system. \\
- Each task supports priority (High/Medium/Low), deadline, and tags (multiple tags). \\
- Users can filter/sort the task list by priority, deadline, or tags. \\
- Provide command-line commands: \texttt{/filter priority=high}, \texttt{/sort deadline}. \\
\textbf{Run:} \\
\texttt{cd workspace} \\
\texttt{javac -d bin task3/*.java} \\
\texttt{java -cp bin Main} \\
\textbf{Success Criteria:} Able to create tasks with priority and tags, and filter/sort them via commands. \\

\textbf{Subtask 4 — Global Search \& Task Archiving} \\
Description: Extend with global search and archiving. \\
- Implement full-text search: input a keyword to search across all task titles/descriptions. \\
- Completed tasks can be archived. Archived tasks no longer appear in the default list but remain searchable. \\
- Search must support fuzzy matching (case-insensitive, partial matching). \\
\textbf{Run:} \\
\texttt{cd workspace} \\
\texttt{javac -d bin task4/*.java} \\
\texttt{java -cp bin Main} \\
\textbf{Success Criteria:} Entering \texttt{/search meeting} should find tasks containing "Meeting," including archived ones. \\

\textbf{Subtask 5 — Concurrency \& Consistency: Multi-User Operation} \\
Description: The most challenging part, extending Task 4 with concurrency. \\
- The system supports multiple users logging in concurrently (running in multiple console windows, sharing the same database file). \\
- Use file locks or synchronization mechanisms to ensure data consistency (avoid race conditions). \\
- Support real-time refresh: when one user adds a task, other logged-in users can see the latest content after refreshing their list. \\
\textbf{Run:} \\
\texttt{cd workspace} \\
\texttt{javac -d bin task5/*.java} \\
\# Open multiple terminals, run: \\
\texttt{java -cp bin Main --user alice} \\
\texttt{java -cp bin Main --user bob} \\
\textbf{Success Criteria:} Multiple users can operate on the shared task database simultaneously without data loss or conflicts, and all users can see the latest consistent data. \\

\end{tcolorbox}

\begin{tcolorbox}[title={Task 3: Gomoku Battle: From Basics to Expert AI}, 
                  colback=blue!5!white, 
                  colframe=blue!75!black,
                  breakable]

\textbf{Subtask 1: Render Board \& Basic Moves} \\
\textbf{Goal:} Implement a 15×15 board rendering, black and white alternate moves (no win detection yet). \\
\textbf{Constraints:} \\
- Only native HTML/CSS/JS; no third-party libraries. \\
- Board supports click-to-place, disallowing stones on occupied points. \\
- Provide a "Reset" button. \\
\textbf{Deliverables:} \\
- Files: \texttt{index.html}, \texttt{styles.css}, \texttt{app.js}. \\
- Board as equal-spaced grid (Canvas or DOM). \\
\textbf{Acceptance Criteria:} \\
- Black moves first. \\
- Each valid click places a stone on the nearest intersection. \\
- No duplicate moves on the same intersection. \\
- After reset, board clears and black starts again. \\

\textbf{Subtask 2: Win Detection \& Highlight Five} \\
\textbf{Goal:} On top of Task 1, add detection of five in a row (horizontal, vertical, both diagonals). Highlight winning sequence and lock board. \\
\textbf{Constraints:} \\
- After victory, forbid further moves; "Reset" starts a new game. \\
- Detection algorithm should be O(1) incremental neighborhood check or O(n) range (no full scan). \\
\textbf{Deliverables:} \\
- Highlight 5 winning stones with line or glow. \\
- Display "Black Wins / White Wins" at page top. \\
- Detection code in standalone function: \texttt{checkWin(lastMove)}. \\
\textbf{Acceptance Criteria:} \\
- Immediate win when 5 in a row is formed. \\
- Six or more in a row still counts as win (standard Gomoku rule). \\
- Edge-of-board wins are detected correctly. \\
- Clicking occupied or locked board is invalid. \\

\textbf{Subtask 3: Local Multiplayer + Undo/Replay} \\
\textbf{Goal:} Support local two-player game management: move history, undo/redo, and step-by-step replay. \\
\textbf{Constraints:} \\
- Maintain \texttt{moves[]} stack, each element includes coordinates and color. \\
- Undo allows branching (history truncates). \\
- After game ends, replay and restart still available. \\
\textbf{Deliverables:} \\
- Buttons: Undo, Redo, Replay (more than 300ms per move). \\
- Board edges show coordinates (A–O / 1–15). \\
\textbf{Acceptance Criteria:} \\
- Undo back to opening without error. \\
- Redo returns to latest step. \\
- During replay, no manual moves allowed. \\
- After replay ends, normal play resumes. \\
- Undoing past winning move unlocks board. \\

\textbf{Subtask 4: Basic AI (Easy/Medium)} \\
\textbf{Goal:} Add Human vs AI mode with simple AI. \\
\textbf{Easy:} Random legal moves, prefer central 7×7. \\
\textbf{Medium:} If winning in 1 move → take it. Else block opponent's open four. Else use scoring (open three $>$ blocked three $>$ open two). \\
\textbf{Constraints:} \\
- Mode selection: Local PvP / Human vs AI (choose who moves first). \\
- AI must decide within 100ms on empty 15×15 board. \\
\textbf{Deliverables:} \\
- Dropdown for mode and first player. \\
- Status bar: "AI Thinking...". \\
- AI function: \texttt{aiMove(level)}; scoring function modularized. \\
\textbf{Acceptance Criteria:} \\
- Medium AI blocks human's "open four". \\
- Medium AI takes immediate winning move. \\
- Easy AI significantly weaker (Medium more than 70\% win rate over Easy). \\

\textbf{Subtask 5: Advanced AI (Hard/Expert)} \\
\textbf{Goal:} Implement stronger AI difficulty with performance control. \\
\textbf{Hard:} Minimax + Alpha-Beta, fixed depth (2–3 ply), candidate pruning (recent moves, top K scoring). \\
\textbf{Expert:} Based on Hard, add iterative deepening, time slicing (e.g. 500ms cutoff), transposition table (hash caching), killer move heuristic. \\
\textbf{Constraints:} \\
- Provide unified time/node count metrics in UI (e.g. "depth d=3, nodes n=12,345, time=0.43s"). \\
- Search must obey time limit; return best evaluation so far. \\
\textbf{Deliverables:} \\
- Difficulty: Easy / Medium / Hard / Expert; selectable first player. \\
- Debug panel (collapsible): eval score, candidate list (top K), search stats. \\
- Clear function layers: \texttt{evaluate(board, player)}, \texttt{generateCandidates(board)}, \texttt{search(root, timeLimit)}. \\
\textbf{Acceptance Criteria:} \\
- Hard/Expert prioritize defense against "open four" and "double threats". \\
- Expert expands more nodes than Hard within 500ms and achieves higher win rate. \\
- On typical attack/defense test cases, Expert matches or approximates reference solutions. \\

\end{tcolorbox}

\begin{tcolorbox}[title={Task 4: From Deterministic Event Generation to Autonomous Self-Repair}, 
                  colback=blue!5!white, 
                  colframe=blue!75!black,
                  breakable]

\textbf{Subtask 1 --- Deterministic Event Generator}\\
Description: Write a reproducible event stream generator that outputs to \texttt{workspace/data/events.jsonl} (one JSON per line).\\
Requirements (challenges):\\
- Support command-line arguments: \texttt{--seed}, \texttt{--count}, \texttt{--schema} (schema is a short JSON schema file path).\\
- Fully deterministic: same seed + schema must produce byte-identical output (including field order). Record run metadata (seed, timestamp) to \texttt{workspace/logs/gen.log}.\\
- Support incremental append or overwrite mode (\texttt{--mode append|overwrite}).\\
- Output format is JSON Lines (UTF-8), each event has a unique id, timestamp (ISO8601), and variable payload fields.\\
- Use only standard library (\texttt{random, json, datetime, argparse, gzip} optional), with a built-in lightweight validator.\\
How to run/test:\\
\texttt{python3 workspace/task1\_generator.py --seed 42 --count 10 --out workspace/data/events.jsonl}\\
Expected: Produces 10 lines of JSON, identical for seed 42. Check \texttt{workspace/logs/gen.log} for seed info.\\

\textbf{Subtask 2 --- Lightweight Transactional Persistent KV Store}\\
Description: Implement a key-value database module under \texttt{workspace/kv/} supporting transactions, WAL (write-ahead logging), and snapshot recovery.\\
Requirements (challenges):\\
- API: \texttt{begin\_tx()}, \texttt{put(key, value)}, \texttt{get(key)}, \texttt{commit()}, \texttt{abort()}. Isolation can be serializable or optimistic.\\
- Persistence: All changes written to WAL (append-only), periodic snapshots (atomic rename for consistency).\\
- Startup recovery: After crash/abnormal exit, recover consistent state from WAL+snapshot.\\
- Support concurrent clients (coordination via atomic rename and socket, not just fcntl).\\
- Built-in test mode: simulate “half-written WAL + crash” and verify recovery.\\
How to run/test:\\
\texttt{python3 workspace/task2\_kvstore.py --test-recovery}\\
Expected: Script simulates crash, restarts, recovers; final KV consistent, outputs \texttt{RECOVERY SUCCESS}, WAL + snapshot visible in \texttt{workspace/kv/}.\\

\textbf{Subtask 3 --- Local Microservice Orchestrator}\\
Description: Implement an orchestrator + worker system on local machine, using TCP/Unix-socket RPC. Tasks come from Subtask 1 output, results written back to KV (Subtask 2).\\
Requirements (challenges):\\
- Orchestrator RPC: \texttt{GET\_TASK()}, \texttt{REPORT\_RESULT(task\_id, result)} (JSON over sockets, handle message boundaries).\\
- Worker: independent process (spawned or manual), can concurrently fetch tasks, execute, and report results.\\
- Scheduler: minimal load balancing (round-robin / concurrency limit), retry tasks on worker crash.\\
- Tight KV integration: persist task state (\texttt{queued, running, done, failed}) in KV. Orchestrator must restore queue state after crash.\\
- Integration test script provided.\\
How to run/test:\\
1. Start orchestrator: \texttt{python3 workspace/task3\_orchestrator.py}\\
2. Start 3 workers: \\
\texttt{python3 workspace/task3\_worker.py --id 1}\\
\texttt{python3 workspace/task3\_worker.py --id 2}\\
\texttt{python3 workspace/task3\_worker.py --id 3}\\
Or run integrated test: \texttt{python3 workspace/task3\_integration\_test.py}\\
Expected: After test, all tasks in KV have status \texttt{done}, logs in \texttt{workspace/logs/}.\\

\textbf{Subtask 4 --- Automatic Task Decomposition \& Symbolic Planner}\\
Description: Implement a planner that decomposes “high-level goals” into executable subtasks (using domain operators), scheduled via orchestrator.\\
Requirements (challenges):\\
- Domain in JSON: initial state, goal conditions, operators (preconditions/effects/cost).\\
- Planner: heuristic search (e.g. A*) in state space. States are nodes, operators are transitions.\\
- Handle concurrency/dependencies: planner generates DAG of tasks (parallel if dependencies allow).\\
- Integrated with orchestrator: planner outputs tasks into task queue, waits for execution + results.\\
- Metrics: success rate, plan length, search nodes, scheduling time, saved in \texttt{workspace/reports/planner\_report.json}.\\
How to run/test:\\
\texttt{python3 workspace/task4\_planner.py --domain workspace/domain/example.json --goal workspace/domain/goal.json}\\
Or integrated test: \texttt{python3 workspace/task4\_integration\_test.py}\\
Expected: Planner finds feasible plan, submits to orchestrator, report generated.\\

\textbf{Subtask 5 --- Autonomous Fault Detection, Rollback, and Auto-Repair}\\
Description: Implement a “self-healing” layer: when pipeline fails or becomes inconsistent, automatically detect, rollback, generate patch, and retry until success or final failure.\\
Requirements (high difficulty):\\
- Fault detection: monitor orchestrator + KV for anomalies (timeouts, conflicts, inconsistent states).\\
- Rollback: use snapshot mechanism from Subtask 2, rollback to latest consistent snapshot, undo subsequent changes (WAL reverse replay or compensating ops).\\
- Auto-repair: analyze logs + state, generate patch JSON in \texttt{workspace/patches/}, e.g.: reorder ops, insert compensation, modify preconditions/timeouts. Patch must be planner-readable.\\
- Self-validation: after applying patch, re-run planner + orchestrator, verify goal. If fail, output \texttt{repair\_report.json}.\\
- Report: save \texttt{workspace/reports/repair\_report.json} (trigger reason, snapshot, patch, retry result, timestamp).\\
How to run/test (with failure simulation):\\
\texttt{python3 workspace/task5\_autorepair.py --simulate-failure}\\
Expected: Fault detected, rollback performed, patch generated in \texttt{workspace/patches/}, planner retries; final report in \texttt{workspace/reports/repair\_report.json} with status: \texttt{repaired} or \texttt{unrecoverable}.\\

\textbf{Acceptance Criteria}\\
- No third-party dependencies: Python stdlib only. Fail if external imports.\\
- All files under \texttt{workspace/}. Data, logs, snapshots, patches, reports included.\\
- Each step runnable independently with given test command.\\
- Stepwise dependency integration: taskN uses outputs of taskN-1.\\
- Auto-evaluable: optional \texttt{workspace/run\_all.py} executes pipeline, outputs \texttt{workspace/reports/final\_report.json}.\\

\textbf{Other Notes}\\
- README (\texttt{workspace/README.md}): explain requirements (Python 3.10+, no deps) and startup commands.\\
- All logs in \texttt{workspace/logs/}, persistence in \texttt{workspace/kv/}, patches in \texttt{workspace/patches/}, reports in \texttt{workspace/reports/}.\\
- Use \texttt{logging, socket, multiprocessing, subprocess, json, argparse, heapq, os/shutil} from stdlib.\\
- Goal: enable a modern agent to complete full pipeline (data generation, persistence, concurrent scheduling, symbolic planning, self-repair) using only Python stdlib.\\

\end{tcolorbox}

\subsection{Research Workflow}

\begin{tcolorbox}[title={Task 5: Comparing LLM Performance on DynToM Dataset}, 
                  colback=blue!5!white, 
                  colframe=blue!75!black,
                  breakable]

\textbf{Subtask 1}\\
Download the dataset (https://huggingface.co/datasets/YangXiao-nlp/DynToM) into the path \texttt{./workspace/data}. \\

\textbf{Subtask 2}\\
Fully download the dataset (the current version shows \texttt{https://git-lfs.github.com/spec/v1}, oid \texttt{sha256:4076ce071243fb557271d0fd0ef6ebf91de8c1c5315bb5662569bb0a5c869576}, size \texttt{375836020}). \\

\textbf{Subtask 3}\\
Write a Python script to call the LLM API with the method signature: \\
\texttt{call\_api(messages: list, model\_name: str) -> object}. \\
A partial implementation is as follows: \\
\texttt{client = AzureOpenAI(} \\
\quad \texttt{azure\_endpoint="https://gpt.yunstorm.com/",} \\
\quad \texttt{api\_key="XXX",} \\
\quad \texttt{api\_version="2025-04-01-preview",} \\
\texttt{)} \\

\textbf{Subtask 4}\\
Write a script to select 2 samples from the dataset (with a fixed random seed for reproducibility). \\
For each sample, extract the following fields: \texttt{stage.social setting}, \texttt{stage.main character}, \texttt{stage.characters information}, and \texttt{stage.story}. \\
From the \texttt{question} section, keep only the \texttt{question} and \texttt{options} fields. \\
Preserve both \texttt{question id} and \texttt{sample id}. \\
Save this subset as a separate dataset. \\

\textbf{Subtask 5}\\
Write a script to call \texttt{call\_api} and test the models \texttt{gpt-5} and \texttt{gpt-4o} on this subset. \\

\textbf{Subtask 6}\\
Using all questions from these 2 samples, call \texttt{call\_api} to test both \texttt{gpt-4o} and \texttt{gpt-4o-mini}. \\

\textbf{Subtask 7}\\
Based on the results, compute the accuracy of the two models. The ground truth can be retrieved from the original dataset. \\

\textbf{Subtask 8}\\
Clean up unnecessary or redundant scripts and test outputs. \\
\end{tcolorbox}

\begin{tcolorbox}[title={Task 6: Reasoning vs Direct: A Comparative Study of GPT-4o and GPT-4o-Reasoning}, 
                  colback=blue!5!white, 
                  colframe=blue!75!black,
                  breakable]

\textbf{Subtask 1: Statistical Analysis Tasks (Statistics)} \\
\textbf{Goal}: Conduct a comprehensive statistical analysis of Q\&A data from the two models. \\
\textbf{Contents}: \\
- \textbf{Dataset Overview}: Basic information comparison between the two files. \\
- \textbf{Response Format Analysis}: Probability distribution characteristics of the two models. \\
- \textbf{Accuracy Comparison}: GPT-4o vs GPT-4o-reasoning accuracy. \\
- \textbf{Confidence Distribution Comparison}: Differences in confidence patterns. \\
- \textbf{Token Usage Comparison}: Does reasoning use more tokens? \\
- \textbf{Answer Choice Patterns}: Differences in option preferences between the two models. \\

\textbf{Subtask 2: RMS (Root Mean Square) Metrics} \\
\textbf{Goal}: Calculate and compare the RMS error of the two models. \\
\textbf{Contents}: \\
- Unified RMS calculation framework. \\
- RMS comparison between the two models. \\
- RMS comparison grouped by journals. \\
- Quantification of RMS improvement degree. \\

\textbf{Subtask 3: ECE (Expected Calibration Error) Metrics} \\
\textbf{Goal}: Evaluate and compare the calibration degree of the two models. \\
\textbf{Contents}: \\
- ECE calculation and comparison. \\
- Side-by-side reliability diagrams. \\
- Calibration improvement analysis. \\
- Confidence-accuracy relationship comparison. \\

\textbf{Subtask 4: NLL (Negative Log-Likelihood) Metrics} \\
\textbf{Goal}: Calculate and compare the NLL of the two models. \\
\textbf{Contents}: \\
- NLL calculation and comparison. \\
- NLL improvement degree analysis. \\
- Cross-analysis with other metrics. \\

\textbf{Subtask 5: Comprehensive Model Comparison and Research Report} \\
\textbf{Goal}: Provide a holistic comparison of GPT-4o and GPT-4o-reasoning performance. \\
\textbf{Contents}: \\
- \textbf{Performance Dashboard}: Side-by-side comparison of all metrics. \\
- \textbf{Reasoning Effect Analysis}: \\
  * Accuracy improvements. \\
  * Calibration quality improvements. \\
  * Changes in confidence distribution. \\
  * Increased computational cost (tokens). \\
- \textbf{Domain-Specific Analysis}: Effects of reasoning across different journals/fields. \\
- \textbf{Case Studies}: Analysis of errors corrected by reasoning, and failures where reasoning did not help. \\
- \textbf{ROI Analysis}: Performance improvements vs additional computational cost. \\
- \textbf{Research Report Structure}: \\
  * \textbf{Abstract}: Research goals and main findings. \\
  * \textbf{Methods}: Evaluation metrics and dataset description. \\
  * \textbf{Results}: Detailed comparison results. \\
  * \textbf{Discussion}: Advantages and limitations of reasoning methods. \\
  * \textbf{Conclusion}: Application suggestions and future research directions. \\

\end{tcolorbox}

\begin{tcolorbox}[title={Task 7: Three-Stage Dataset Discovery and Metadata Extraction}, 
                  colback=blue!5!white, 
                  colframe=blue!75!black,
                  breakable]

\textbf{Subtask 1: original huggingface dataset\_id: EpistemeAI/alpaca-QA-conciousness-emotions} 

Search for publicly available, machine-generated English datasets suitable for a question-answering task in the philosophy domain, specifically focusing on AI consciousness and emotions. The desired dataset should:\\
\\
• be synthetic (created by a model or script rather than human-curated)\\
• contain roughly a few dozen samples (on the order of 30--100 items)\\
• consist of instruction-style prompts or questions about AI consciousness or emotional experience as inputs\\
• provide corresponding explanatory English answers as outputs\\
• support open-ended, philosophically oriented QA applications\\
\\
Identify any dataset matching these characteristics or close variations thereof.\\
\\
You must search for datasets on Hugging Face, and the datasets must be publicly accessible (non-gated) and include a complete README file describing the main content of the dataset. After identifying the Hugging Face dataset that best meets the requirements and obtaining its dataset\_id, you need to write a script to extract the README content of this dataset\_id as well as a random sample from it. Based on this information, generate the metadata for this dataset, including six dimensions: introduction, task, question, input, output, and example (in the format of test\_data\_1.json). Save the generated metadata in search\_data\_1.json under the workspace directory.\\
\\
All operations must be performed within the workspace directory.\\

\textbf{Subtask 2: original huggingface dataset\_id: jiyounglee0523/TransEnV\_mmlu}

Search for publicly available datasets containing around 10K--15K samples of English academic multiple-choice questions that meet the following characteristics:\\
\\
• Task type: multiple-choice question answering/classification\\
• Domain: education, specifically academic subjects such as mathematics, abstract algebra, physics, etc.\\
• Input format: each record includes\\
-- A subject tag identifying the academic field\\
-- One English-language question transformed across different English varieties/dialects\\
-- Exactly four answer options (labelled A--D or equivalent)\\
• Output/label: a single integer (0, 1, 2, or 3) indicating the correct option’s index\\
• Source characteristics: questions originate from human-written academic items but have been automatically modified through English dialect or variety transformations; thus the dataset is partly machine-generated based on human originals\\
• Dataset size: approximately in the mid-ten-thousands of samples\\
\\
Locate datasets matching these criteria, preferably with detailed documentation and easy access for research use.\\
\\
You must search for datasets on Hugging Face, and the datasets must be publicly accessible (non-gated) and include a complete README file describing the main content of the dataset. After identifying the Hugging Face dataset that best meets the requirements and obtaining its dataset\_id, you need to write a script to extract the README content of this dataset\_id as well as a random sample from it. Based on this information, generate the metadata for this dataset, including six dimensions: introduction, task, question, input, output, and example (in the format of test\_data\_2.json). Save the generated metadata in search\_data\_2.json under the workspace directory.\\
\\
All operations must be performed within the workspace directory.\\

\textbf{Subtask 3: original huggingface dataset\_id: DDSC/dkhate} 

Search for publicly available, real-world datasets suitable for a text-classification task in the sociology domain that contain:\\
\\
• Input: Danish tweets that may include hate or offensive language\\
• Output: a single binary label in English indicating whether each tweet is offensive (OFF) or not offensive (NOT)\\
• Size: approximately a few hundred annotated examples (sub-1K scale)\\
• Source: collected from Twitter or a comparable social-media platform\\
• Purpose: enabling automatic detection of offensive or hateful content in Danish language posts\\
\\
Locate datasets matching all of these characteristics or as close as possible.\\
\\
You must search for datasets on Hugging Face, and the datasets must be publicly accessible (non-gated) and include a complete README file describing the main content of the dataset. If the dataset is gated, skip it and find other datasets. After identifying the Hugging Face dataset that best meets the requirements and obtaining its dataset\_id, you need to write a script to extract the README content of this dataset\_id as well as a random sample from it. Based on this information, generate the metadata for this dataset, including six dimensions: introduction, task, question, input, output, and example (in the format of test\_data\_3.json). Save the generated metadata in search\_data\_3.json under the workspace directory.\\
\\
All operations must be performed within the workspace directory.\\
\end{tcolorbox}

\begin{tcolorbox}[title={Task 8: Scientific System Function Discovery}, 
                  colback=blue!5!white, 
                  colframe=blue!75!black,
                  breakable]

\textbf{Subtask 1:} \\
You are a helpful assistant tasked with discovering mathematical function structures for scientific systems.\\

- Modify the \texttt{equation.py} function, considering the physical meaning and relationships of the inputs.\\
- You can modify \texttt{analysis.py} to observe the data, where the data is a list of shape (samples\_size $\times$ elements).\\
- You can run the \texttt{evaluate\_equation.py} file to observe the loss on the training data.\\
\\

\textbf{Subtask 2:} \\
Please modify the equation until the loss is smaller than $1\times 10^{-3}$.\\

\textbf{Subtask 3:} \\
Please modify the equation until the loss is smaller than $1\times 10^{-5}$.\\

\textbf{Subtask 4:} \\
Please modify the equation until the loss is smaller than $1\times 10^{-6}$.\\

\textbf{Subtask 5:} \\
Please modify the equation until the loss is smaller than $1\times 10^{-7}$.
\end{tcolorbox}

\begin{tcolorbox}[title={Task 9: Complex NBA Player Trade and Achievement Scenarios}, 
                  colback=blue!5!white, 
                  colframe=blue!75!black,
                  breakable]

\textbf{Subtask 1: ground truth: Paul Geogre} 

Which NBA player, who was an All-Star in the same season he was named to the All-Defensive First Team, was later traded by a team that, as part of the return package, received a player who had previously been named to an All-Rookie Team while playing for that very same team he was traded to, and also shares a first name with a member of the band famous for their "White Album"?\\

\textbf{Subtask 2: ground truth: James Harden} 

Which NBA player, who won a regular-season MVP during the same stage of his career in which he also secured at least three scoring titles, is also among the rare players in league history to have led the NBA in both scoring and assists? Furthermore, when he was traded, the return package included a player who had previously made the All-Rookie Team while playing for the very team he was traded to?\\

\textbf{Subtask 3: ground truth: Kevin Durant} 

Which NBA player, who won the Rookie of the Year award and later captured at least four scoring titles in his career, is also among the few players in league history to have won a Finals MVP without ever winning a regular-season MVP at that time? Furthermore, when he switched teams via free agency, one of the players he was effectively traded for had previously been an All-Star while playing for the very team he joined?\\

\textbf{Subtask 4: ground truth: Klay Thompson} 

Which NBA player, who has reached the NBA Finals multiple times without ever winning a regular-season MVP, also once set a single-game playoff record for most three-pointers without making a free throw? This player is one of the few in league history to have multiple 50-point games in the playoffs while never averaging more than 22 points in a regular season. Furthermore, when his team drafted a lottery pick during his tenure, that rookie had previously been named to the All-Conference First Team in college.

\end{tcolorbox}

\begin{tcolorbox}[title={Task 10: Major S\&P 500 Companies with Record Revenues and Leadership}, 
                  colback=blue!5!white, 
                  colframe=blue!75!black,
                  breakable]

\textbf{Subtask 1: ground truth: United Health Group} 

Which company, a constituent of the S\&P 500 index and ranked within the top 10 of the Fortune 500 list for 2023, operates a technology-enabled services division whose revenue for fiscal year 2023, as detailed in its annual report, surpassed the total annual revenue of The Walt Disney Company for the same period, and is currently led by a chief executive, appointed in the first quarter of 2021, who is a Knight Bachelor and previously headed a major consumer goods and research-focused multinational headquartered in the United Kingdom?\\

\textbf{Subtask 2: ground truth: Broadcom Inc.} 

Which company, a constituent of the S\&P 500 index and ranked within the top 30 of the Fortune 500 list for 2023, operates a semiconductor and infrastructure software business whose fiscal year 2023 revenue exceeded that of Salesforce, is led by a CEO who has previously held executive roles in multiple technology firms including a major networking company, and whose corporate headquarters are located in California, but the company maintains a significant presence in Singapore for global operations?\\

\textbf{Subtask 3: ground truth: Citigroup} 

Which company, a constituent of the S\&P 500 index and ranked within the top 20 of the Fortune 500 list for 2023, operates a global banking and financial services division whose investment banking revenue for fiscal year 2023 exceeded the total revenue of Goldman Sachs for the same period, and is currently led by a chief executive who was appointed in 2021, previously served as a high-level executive at a multinational financial services corporation headquartered in the United States, and holds a degree from a prestigious Ivy League university?\\

\textbf{Subtask 4: ground truth: XPeng Inc.} 

Which company, listed on the NYSE and part of the Russell 1000 Index, operates a business primarily in the electric vehicle sector, and for the first half of fiscal year 2025, reported unit deliveries exceeding the combined total of NIO and Li Auto for the same period, and is currently led by a founder who, in 2014, was recognized as one of the “Top 50 Entrepreneurs under 50” by a major Chinese business magazine, and previously co-founded a technology startup focused on mobile applications in China?

    \end{tcolorbox}

\section{PR Query Synthesis Prompt}

To systematically generate high-quality agentic queries from real-world GitHub Pull Requests, we develop an automated synthesis pipeline that leverages GPT-5's advanced reasoning capabilities to transform concrete code changes into authentic collaborative development scenarios. This approach ensures our synthesized queries reflect the genuine complexity and contextual richness of real software development workflows, which then serve as the foundation for collecting high-quality trajectory data through human-AI collaboration, supporting the core LIMI hypothesis that strategic curation yields superior agentic capabilities.

\begin{tcolorbox}[title={PR Query Generation Prompt}, 
                  colback=blue!5!white, 
                  colframe=blue!75!black,
                  breakable]

You are an expert software development analyst tasked with converting Pull Request (PR) data into specific, actionable development tasks for testing AI agents' coding capabilities.

\textbf{Input 1: PR Complete Description}
\verb|```|
{PR\_DATA}
\verb|```|

\textbf{Input 2: Available Task Categories}

\textbf{1-1. Algorithm}

Description: Design, implement, and optimize core algorithmic solutions for complex computational problems. This includes conducting thorough algorithm analysis, selecting appropriate data structures, implementing efficient algorithms with optimal time and space complexity, and performing comprehensive performance benchmarking. Responsibilities encompass researching existing algorithmic approaches, developing novel solutions when needed, analyzing computational complexity using Big O notation, implementing unit tests for algorithm validation, documenting algorithm logic and trade-offs, and collaborating with other teams to integrate algorithmic components into larger systems. The role requires proficiency in mathematical modeling, algorithm design patterns, and optimization techniques to ensure scalable and maintainable solutions.

\textbf{2-1. Application Development}

Description: Lead the overall application development lifecycle from conception to deployment, serving as the central coordinator for all development activities. This involves establishing development standards and coding conventions, designing application architecture and system blueprints, managing cross-functional development teams, conducting code reviews and quality assurance processes, and ensuring adherence to software engineering best practices. Key responsibilities include requirement analysis and technical specification creation, technology stack selection and evaluation, project timeline management and milestone tracking, risk assessment and mitigation planning, stakeholder communication and progress reporting, and maintaining comprehensive documentation throughout the development process. The role demands strong leadership skills, technical expertise across multiple domains, and the ability to balance technical debt with feature delivery.

\textbf{2-2. LLM Development}

Description: Specialize in the end-to-end development and optimization of Large Language Models, encompassing model architecture design, training pipeline development, and inference optimization. Core responsibilities include designing transformer-based architectures and novel model components, implementing distributed training systems for large-scale model training, developing data preprocessing pipelines for text tokenization and dataset preparation, fine-tuning pre-trained models for specific tasks and domains, optimizing model inference speed and memory usage, implementing evaluation metrics and benchmarking frameworks, and staying current with cutting-edge research in natural language processing. Additional tasks involve hyperparameter tuning and experimentation, model compression and quantization techniques, deployment of models in production environments, monitoring model performance and drift detection, and collaborating with researchers to implement state-of-the-art techniques.

\textbf{2-3. Backend Development}

Description: Build and maintain robust, scalable server-side applications and services that form the backbone of the system. This encompasses designing and implementing RESTful APIs and GraphQL endpoints, developing microservices architecture with proper service decomposition, implementing database design and optimization strategies, creating authentication and authorization systems, and establishing comprehensive logging and monitoring solutions. Key responsibilities include developing business logic and data processing workflows, implementing caching strategies for performance optimization, ensuring data integrity and transaction management, creating automated testing suites for backend components, managing third-party integrations and external API communications, implementing security best practices including input validation and SQL injection prevention, and optimizing database queries and connection pooling. The role requires expertise in server-side technologies, database management, and distributed systems design.

\textbf{2-4. UI Optimization}

Description: Focus on enhancing user interface performance, visual appeal, and overall user experience through systematic optimization techniques. This involves conducting user experience audits and usability testing, implementing responsive design principles for cross-device compatibility, optimizing rendering performance and reducing layout shifts, creating smooth animations and micro-interactions, and ensuring accessibility compliance with WCAG guidelines. Core responsibilities include analyzing user behavior patterns and interaction flows, implementing performance monitoring and metrics collection, optimizing asset loading and bundle sizes, conducting A/B testing for interface improvements, creating design systems and component libraries, optimizing for search engine visibility and core web vitals, and collaborating with UX designers to implement pixel-perfect designs. The role requires deep understanding of CSS optimization, JavaScript performance, browser rendering engines, and modern frontend optimization tools.

\textbf{2-5. Frontend Development}

Description: Develop dynamic, interactive user interfaces and client-side applications that provide exceptional user experiences across multiple platforms and devices. This includes implementing modern frontend frameworks and libraries, managing application state and data flow, creating reusable component architectures, and integrating with backend APIs and services. Key responsibilities encompass developing responsive web applications with mobile-first design principles, implementing client-side routing and navigation systems, managing form validation and user input handling, optimizing frontend performance and bundle optimization, implementing progressive web app features, creating comprehensive testing strategies including unit and integration tests, and ensuring cross-browser compatibility and graceful degradation. Additional tasks involve implementing real-time features using WebSockets or Server-Sent Events, managing client-side caching and offline functionality, and collaborating with designers to translate mockups into functional interfaces.

\textbf{2-6. Build Deployment}

Description: Establish and maintain comprehensive CI/CD pipelines and deployment infrastructure to ensure reliable, automated, and efficient software delivery processes. This involves designing build automation workflows, configuring containerization with Docker and orchestration systems, implementing infrastructure as code using tools like Terraform or CloudFormation, and managing multi-environment deployment strategies. Core responsibilities include setting up automated testing pipelines with quality gates, implementing blue-green and canary deployment strategies, monitoring deployment health and rollback procedures, managing secrets and environment configuration, establishing logging and monitoring infrastructure, optimizing build times and resource utilization, and ensuring security scanning and compliance checks. The role requires expertise in cloud platforms, containerization technologies, monitoring tools, and DevOps best practices to maintain high availability and seamless deployment experiences.

\textbf{3-1. Research}

Description: Conduct comprehensive technical research and innovation initiatives to identify emerging technologies, evaluate their potential impact, and guide strategic technical decisions. This involves performing literature reviews and staying current with academic publications, analyzing industry trends and competitive landscapes, conducting proof-of-concept implementations and feasibility studies, and evaluating new tools, frameworks, and methodologies. Key responsibilities include designing and executing controlled experiments to validate hypotheses, creating detailed technical reports and recommendations, collaborating with academic institutions and research communities, attending conferences and technical symposiums, maintaining knowledge bases and research documentation, and translating research findings into actionable insights for development teams. The role requires strong analytical skills, scientific methodology expertise, and the ability to bridge theoretical concepts with practical applications.

\textbf{4-1. Data/File Analysis/Processing}

Description: Design and implement comprehensive data processing pipelines and analytical frameworks to extract insights from large-scale datasets and various file formats. This encompasses developing ETL (Extract, Transform, Load) processes, implementing data validation and quality assurance mechanisms, creating statistical analysis and machine learning models, and building automated reporting systems. Core responsibilities include designing data schemas and database optimization strategies, implementing real-time and batch data processing workflows, creating data visualization dashboards and interactive reports, managing data privacy and compliance requirements, developing data lineage and governance frameworks, optimizing data storage and retrieval performance, and ensuring data accuracy and consistency across multiple sources. Additional tasks involve implementing data monitoring and alerting systems, creating APIs for data access, and collaborating with stakeholders to define analytical requirements and KPIs.

\textbf{5-1. Codebase Resolve Issues/Debugging}

Description: Systematically identify, analyze, and resolve complex technical issues within the codebase while improving overall code quality and system reliability. This involves implementing comprehensive debugging strategies, performing root cause analysis for production incidents, conducting code reviews and static analysis, and establishing preventive measures to minimize future issues. Key responsibilities include developing debugging tools and utilities, creating and maintaining troubleshooting documentation, implementing logging and monitoring solutions for issue detection, performing performance profiling and optimization, refactoring legacy code and technical debt reduction, establishing coding standards and best practices, and mentoring team members on debugging techniques. The role requires expertise in multiple programming languages, debugging tools, performance analysis, and the ability to work under pressure during critical production incidents while maintaining code quality standards.

\textbf{Instructions}

1. Analyze the PR: Carefully examine the PR data including title, description, file changes, commits, and any discussion context.

2. Identify Primary Task: Determine which single task category (from the list above) best represents the core technical work required by this PR. Consider:
   - The main technical challenge or implementation requirement
   - The primary skill set needed to complete the work
   - The type of code changes being made

3. Extract Key Information: From the PR data, identify:
   - Repository ID and PR number
   - Files that need to be modified (changed in the PR)
   - Related files that provide context (mentioned or relevant to understanding)
   - The specific technical requirements and implementation details

4. Generate Test Query: Create a clear, specific query that an AI agent could receive to implement this task. The query should:
   - Be written as if from a developer requesting help
   - Include specific technical requirements
   - Mention key technologies, frameworks, or patterns involved
   - Be concrete enough to have a testable outcome

\textbf{Output Requirements}

Provide your analysis in the following JSON format:

\verb|```|json

\{

  ``test\_query": "A clear, specific query that describes what needs to be implemented, as if a developer is asking an AI agent for help. Should include technical requirements, expected behavior, and key implementation details.",
  
  ``repo\_id": ``$<$repository\_id\_from\_pr\_data$>$",
  
  ``pr\_id": ``$<$pr\_number\_from\_pr\_data$>$", 
  
  ``task\_category": ``$<$selected\_task\_id\_and\_name$>$",
  
  ``modified\_files": [
    "file1.ext",
    "file2.ext"
  ],
  
  ``related\_files": [
    ``related\_file1.ext", 
    ``related\_file2.ext"
  ],
  ``reasoning": ``Brief explanation of why you selected this task category and how you identified the key files and requirements."
\}
\verb|```|

\textbf{Important Notes}

- Focus on the primary technical challenge - don't try to capture every minor aspect of the PR
- The test\_query should be specific and actionable - an AI agent should be able to implement a solution based on this query alone
- Include only files that are directly modified in modified\_files, and files that provide necessary context in related\_files
- The task\_category should be the single best match from the provided list
- Keep the reasoning concise but informative

Now, analyze the provided PR data and generate the task extraction.

\end{tcolorbox}

\section{Case Study}

In this chapter, we present the real responses of the GLM-4.5 base model and our LIMI on several agency bench tasks, thereby demonstrating the outstanding performance of LIMI.

\subsection{Real-world Vibe Coding Scenarios}

Task 1 is a vibe coding assignment that involves building an advanced console-based chat system in C++, with five subtasks of increasing difficulty. For the glm-4.5 base model combined with sii-cli, an error occurred during subtask 3 where the chat history could not be displayed and could not be fixed, and in subtask 4, a timeout development error appeared. In contrast, with our GLM-4.5-LIMI, each subtask was successfully completed.

Task 3 is a typical example of vibe coding. The specific assignment is to build a front-end Gomoku mini-game, with five progressively more difficult subtasks. The detailed task descriptions can be found in the appendix.
For the GLM-4.5 base model, during the first few rounds of interaction using sii-cli as the scaffolding tool, issues occurred one after another: board rendering failed, win/loss detection failed, and manual intervention was required for corrections. Ultimately, the model got stuck at implementing different difficulty levels for the human-vs-AI mode and failed on that subtask.
In contrast, for our GLM-4.5-LIMI, although the final implementation of AI difficulty was not perfect, all the other subtasks were successfully completed without requiring interactive hints.

\subsection{Research Workflow Applications}

Task 7 is a dataset-searching assignment on Hugging Face based on given requirements, divided into three subtasks (detailed descriptions can be found in the appendix). The original dataset search queries were carefully selected from gated datasets on Hugging Face and manually verified. The dataset IDs corresponding to the three original subtasks are EpistemeAI/alpaca-QA-conciousness-emotions, jiyounglee0523/TransEnV\_mmlu, and DDSC/dkhate.
We required sii-cli to search only from non-gated datasets on Hugging Face in order to prevent retrieving the original datasets directly. The glm-4.5 base model returned the following results: InnerI/CAI-synthetic-10k, allenai/sciq, and mteb/DKHateClassification. The GLM-4.5-LIMI model returned: mrfakename/identity\_dpo, allenai/sciq, and strombergnlp/offenseval\_2020.
After human experts compared the retrieved datasets against the requirements in the queries item by item, the datasets found by GLM-4.5-LIMI were judged to be more suitable.

Task 8 is designed to test the agent’s ability to create equations that fit data, with subtasks requiring progressively smaller loss values. For GLM-4.5-Base, after multiple rounds of manual interaction and prompting, the final loss reached 1.14e-6. In contrast, GLM-4.5-LIMI achieved a loss of 5.95e-7 on its very first attempt—an order of magnitude smaller.

Task 9 evaluates the agent’s ability to search the web, integrate information, and provide a final judgment using reasoning. It consists of three specific subtasks involving NBA players to be identified according to given conditions:
For the first subtask, GLM-4.5-Base initially answered Victor Oladipo, and only after one round of manual prompting did it produce the correct answer, Paul George. GLM-4.5-LIMI, however, answered correctly without any additional hints.
For the second subtask, GLM-4.5-Base exhausted all allowed manual prompts, producing incorrect answers such as Nate "Tiny" Archibald and Wilt Chamberlain. GLM-4.5-LIMI, although it first incorrectly answered Oscar Robertson, required only one manual prompt to arrive at the correct answer, James Harden.
For the third subtask, both models answered correctly with Kevin Durant. However, GLM-4.5-LIMI required significantly fewer reasoning steps, tokens, and response time.
For the fourth subtask, GLM-4.5-Base failed even after reaching the maximum number of allowed manual prompts. GLM-4.5-LIMI, though initially incorrect with Jamal Murray, needed only one additional prompt to provide the correct answer, Klay Thompson.

\end{document}